\documentclass[runningheads]{llncs}

\usepackage{multirow}
\usepackage[table,xcdraw,dvipsnames]{xcolor}
 
\usepackage{eccv}

\usepackage{eccvabbrv}

\usepackage{graphicx}
\usepackage{booktabs}

\usepackage{bm}
\usepackage{mathtools}
\usepackage{bbm}
\usepackage{nicefrac}
\DeclarePairedDelimiter\abs{\lvert}{\rvert}

\newcommand{\fraccomma}{\genfrac{}{}{0pt}{}{}{,}}
\newcommand{\defeq}{\vcentcolon=}

\newcommand*{\bfPar}[1]{{\scriptsize $\ $ \par}\noindent\textbf{#1 }}

\usepackage[accsupp]{axessibility}  % Improves PDF readability for those with disabilities.

\usepackage{hyperref}
\hypersetup{pagebackref,breaklinks,colorlinks,citecolor=eccvblue}

\usepackage{orcidlink}

\begin{document}

\title{Deblur \textit{e}-NeRF: NeRF from Motion-Blurred Events under High-speed or Low-light Conditions}

\titlerunning{Deblur \textit{e}-NeRF}

\author{Weng Fei Low\orcidlink{0000-0001-7022-5713}\index{Low, Weng Fei} \and
Gim Hee Lee\orcidlink{0000-0002-1583-0475}\index{Lee, Gim Hee}}

\authorrunning{W. F. Low and G. H. Lee}

\institute{
The NUS Graduate School's Integrative Sciences and Engineering Programme (ISEP)\\
Institute of Data Science (IDS), National University of Singapore\\
Department of Computer Science, National University of Singapore\\
\email{\{wengfei.low, gimhee.lee\}@comp.nus.edu.sg}\\
\url{https://wengflow.github.io/deblur-e-nerf}
}

\maketitle

\begin{abstract}
The distinctive design philosophy of event cameras makes them ideal for high-speed, high dynamic range \& low-light environments, where standard cameras underperform. However, event cameras also suffer from motion blur, especially under these challenging conditions, contrary to what most think. This is due to the limited bandwidth of the event sensor pixel, which is mostly proportional to the light intensity. Thus, to ensure event cameras can truly excel in such conditions where it has an edge over standard cameras, event motion blur must be accounted for in downstream tasks, especially reconstruction. However, no prior work on reconstructing Neural Radiance Fields (NeRFs) from events, nor event simulators, have considered the full effects of event motion blur. To this end, we propose, Deblur \textit{e}-NeRF, a novel method to directly and effectively reconstruct blur-minimal NeRFs from motion-blurred events, generated under high-speed or low-light conditions. The core component of this work is a physically-accurate pixel bandwidth model that accounts for event motion blur. We also introduce a threshold-normalized total variation loss to better regularize large textureless patches. Experiments on real \& novel realistically simulated sequences verify our effectiveness. Our code, event simulator and synthetic event dataset are open-sourced.

  \keywords{Neural Radiance Field \and Motion blur \and Event camera}
\end{abstract}

\begin{figure}[t]
    \centering
    \includegraphics[width=1.0\linewidth]{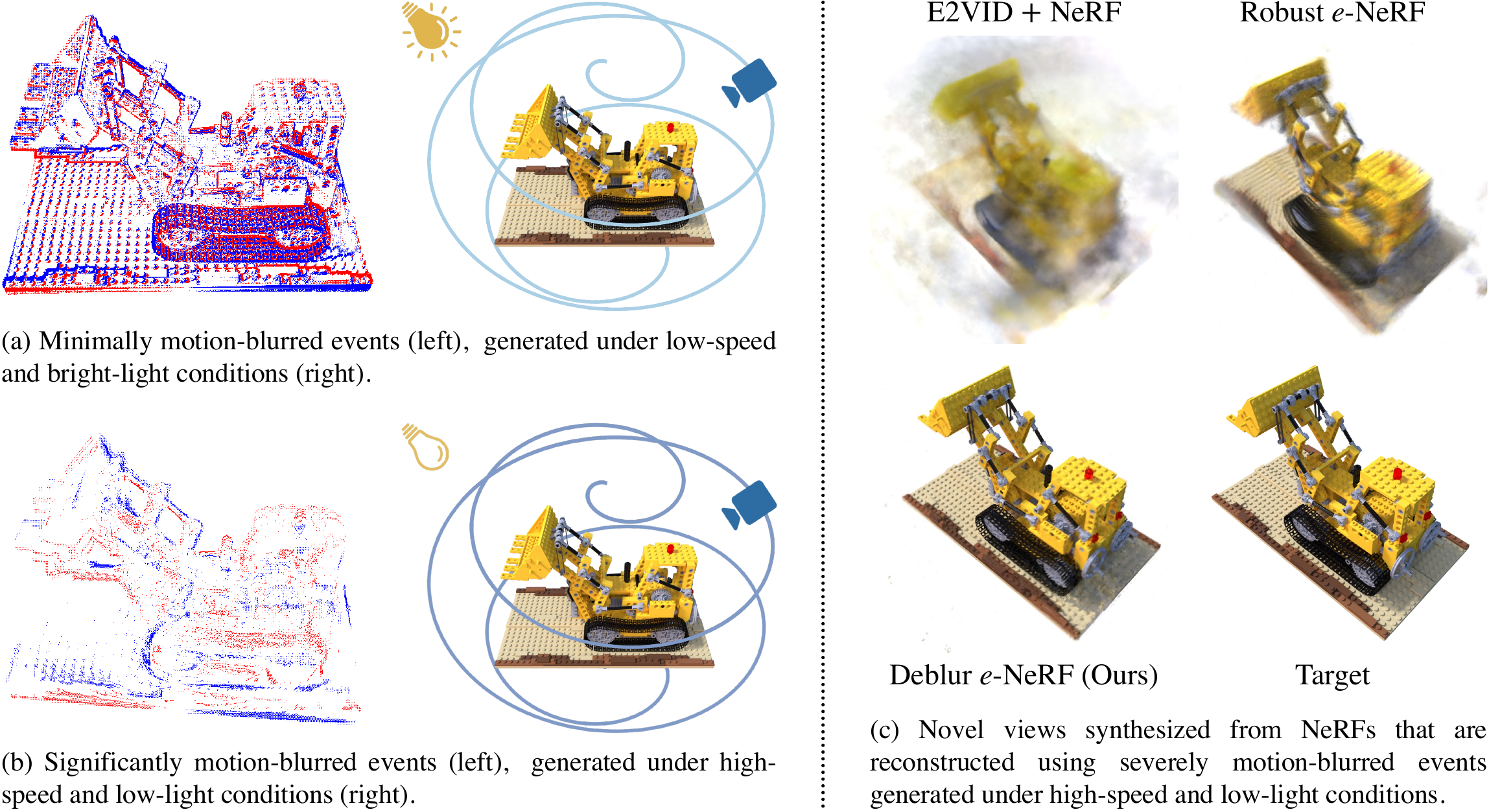}
    
    \caption{Existing works on NeRF reconstruction from moving event cameras heavily rely on (a). In contrast, Deblur \textit{e}-NeRF is able to directly and effectively reconstruct blur-minimal NeRFs from (b), as shown in (c).}
    \label{fig:teaser}
\end{figure}

\section{Introduction}
\label{sec:intro}

Event cameras offer a complementary approach to visual sensing, commonly achieved with frame-based cameras. Instead of capturing intensity images at a fixed rate, event cameras asynchronously detect changes in log-intensity per pixel and output a stream of \textit{events}, each encoding the time instant, pixel location and direction of change. Such a stark contrast in design philosophy enable event cameras to offer many attractive properties over standard cameras, \eg high dynamic range, high temporal resolution, low latency and low power \cite{gallego2020_event_survey}.

These desirable properties make event cameras particularly ideal for applications that involve high-speed motion, \textit{High Dynamic Range} (HDR)/low-light scenes and/or a strict power budget, such as in robotics, augmented reality, surveillance and mobile imaging. Since these are exactly the operating conditions where standard cameras underperform, event cameras meaningfully complements standard cameras. This is clearly demonstrated in the recent success of image deblurring \cite{zhang2022_unifying_motion,xu2021_motion_deblurring,jiang2020_learning_event,lin2020_learning_event_driven}, attributed to the addition of an event camera.

\begin{figure}[t]
    \centering
    \includegraphics[width=1.0\linewidth]{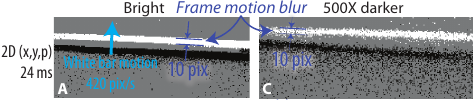}
    
    \caption{Event motion blur from a white bar moving on a black background (From \cite{hu2021_v2e})}
    \label{fig:motion_blur}
\end{figure}

Nonetheless, event cameras also suffer from motion blur \cite{hu2021_v2e,yang2024_latency,liu2024_nernet}, especially under high speed or low light, albeit much less severe than standard cameras. This is contrary to what most think, as it has not been widely documented and discussed in the computer vision community. In general, motion blur of events are manifested as a time-varying latency on the event generation process. In severe cases, a significant ``loss or introduction of events'' may also occur, especially the former. This leads to artifacts such as event trails and blurring of edges, when visualizing events in 2D as their image-plane projection, as shown in \cref{fig:teaser,fig:motion_blur}. Event motion blur can be attributed to the limited bandwidth of the event sensor pixel, which is mostly proportional to the incident light intensity \cite{lichtsteiner2006_aer,mcreynolds2022_experimental, delbruck1995_phototransduction,delbruck1993_thesis,graca2021_unraveling, graca2023_optimal_biasing,hu2021_v2e}. It bounds the minimum event detection latency, maximum frequency of change detectable and hence maximum event generation rate.

Therefore, to ensure event cameras can truly excel under conditions of high-speed or low-light, where it has an edge over standard cameras, it is crucial to account for event motion blur in downstream tasks, especially reconstruction. While recent works on reconstructing \textit{Neural Radiance Fields} (NeRFs) \cite{mildenhall2020_nerf}  from events \cite{low2023_robust-e-nerf, rudnev2022_eventnerf, hwang2022_evnerf, klenk2022_enerf}, and possibly images \cite{klenk2022_enerf, qi2023_e2nerf, ma2023_de-nerf, cannici2024_ev_deblurnerf}, have shown impressive results, none of them accounted for event motion blur, which limits their performance. Moreover, none of the existing event simulators \cite{rebecq2018_esim, hu2021_v2e, joubert2021_icns} model the full non-linear behaviour of the pixel bandwidth under arbitrary lighting conditions. They at most model the 1\textsuperscript{st}-order behavior exhibited under extreme low light.

\bfPar{Contributions.}
We propose Deblur \textit{e}-NeRF, a novel method to directly and effectively reconstruct blur-minimal NeRFs from motion-blurred events, generated under high-speed motion or low-light conditions.

Specifically, we introduce a physically-accurate pixel bandwidth model to account for event motion blur under arbitrary speed \& lighting conditions. We also present a discrete-time variant of the model, a numerical solution to its transient response \& an importance sampling strategy, to enable its computational implementation. We incorporate them as part of the event generation model to reconstruct blur-minimal NeRFs via \textit{Analysis-by-Synthesis} \cite{kato2020_dr_survey}, which also supports the joint optimization of unknown pixel bandwidth model parameters. We also introduce a novel \textit{threshold-normalized total variation loss} to better regularize large textureless patches in the scene. Ambiguities in the reconstruction are resolved by performing the proposed \textit{translated-gamma correction}, which takes the pixel bandwidth model into consideration. Experiments on new event sequences, simulated with an improved ESIM \cite{rebecq2018_esim,low2023_robust-e-nerf} using our pixel bandwidth model, \& real sequences from EDS \cite{hidalgo2022_eds}, clearly validate the effectiveness of Deblur \textit{e}-NeRF. Our code, event simulator \& synthetic event dataset are open-sourced.

\section{Related Work}
\label{sec:rel_work}

\bfPar{Image Motion Deblurring.}
Image motion blur can be simply modeled as an average of incident light intensity over the exposure time of the image \cite{potmesil1983_motion_blur}, which is a unity-gain \textit{Linear Time-Invariant} (LTI) \textit{Low-Pass Filter} (LPF). Thus, the severity of image motion blur is invariant to lighting, unlike event motion blur.

This model may be used in an \textit{Analysis-by-Synthesis} framework \cite{kato2020_dr_survey} to deblur images \cite{park2017_joint_est}. However, a simplified model of spatially filtering images with a time-varying blur kernel \cite{potmesil1983_motion_blur} is more commonly used for this task \cite{shan2008_high_quality,perrone2014_total_variation,krishnan2011_blind_deconv,xu2013_unnatural_l0,cho2009_fast_motion_deblurring,krishnan2009_fast_image_deconv,xu2010_two_phase}. With the rise of deep learning, state-of-the-art image/video deblurring methods \cite{zamir2021_mprnet,son2021_pvdnet,tao2018_srn-deblurnet,zamir2022_restomer,wang2022_uformer} are generally deep image-to-image translation networks.

Despite the rich literature on image motion deblurring, there are no known methods for effective event motion deblurring. While concurrent work \cite{yang2024_latency} proposed to correct event timestamps for motion blur-induced latency (\cref{sec:intro}), the method cannot handle any blur-induced ``loss or introduction of events'' and does not generalize to different cameras and bias settings. Similar generalization issues also plague concurrent work \cite{liu2024_nernet} on night-time events-to-video reconstruction.

\bfPar{NeRF from Motion-Blurred Images.}
A naïve way to reconstruct a blur-minimal \textit{Neural Radiance Field} (NeRF) \cite{mildenhall2020_nerf} from motion-blurred images is to first deblur them using an existing method. Recent works have shown superior performance by integrating either the full \cite{wang2023_bad-nerf,lee2023_exblurf} or simplified \cite{ma2022_deblur-nerf,lee2023_dp-nerf} image motion blur model into an Analysis-by-Synthesis framework, where both NeRF and blur model parameters are jointly optimized, similar to our work. However, there are no known works on NeRF reconstruction from motion-blurred events.

\section{Preliminaries: Robust \textit{e}-NeRF}
\label{sec:prelims}

Robust \textit{e}-NeRF \cite{low2023_robust-e-nerf} is the state-of-the-art for reconstructing NeRFs with event cameras, particularly from temporally sparse and noisy events generated under non-uniform motion. The method consists of 2 key components: a realistic event generation model and a pair of normalized reconstruction losses. After training, NeRF renders are \textit{gamma corrected} to resolve ambiguities in the reconstruction.

\bfPar{Event Generation Model.}
An \textit{Event}, denoted as $\bm{e} = \left( \bm{u}, p, t_{\mathit{prev}}, t_{\mathit{curr}} \right)$, with polarity $p \in \{ -1, +1 \}$ is generated at timestamp $t_{\mathit{curr}}$ when the change in incident log-radiance $\log L$ at a pixel $\bm{u}$, measured relative to a reference value at timestamp $t_{\mathit{ref}}$, shares the same sign as $p$ and possesses a magnitude given by the \textit{Contrast Threshold} associated to polarity $p$, $C_p$. Following the generation of an event, the pixel will be momentarily deactivated for an amount of time determined by the \textit{Refractory Period} $\tau$ and then reset at the end. This event generation model, as illustrated in \cref{fig:egm}, can be succinctly described by:

\begin{equation}
	\Delta \log L \defeq \log L (\bm{u}, t_{\mathit{curr}}) - \log L (\bm{u}. t_{\mathit{ref}}) = pC_p \ , \text{where } t_{\mathit{ref}} = t_{\mathit{prev}} + \tau \ .
    \label{eq:egm}
\end{equation}

\bfPar{Training.}
To reconstruct a NeRF from an \textit{Event Stream} $\mathcal{E} = \{ \bm{e} \}$, provided by a calibrated event camera with known trajectory, a batch of events $\mathcal{E}_{\mathit{batch}}$ is sampled randomly from $\mathcal{E}$ for optimization of the following total training loss:
\begin{equation}
	\mathcal{L} = \frac{1}{\abs{\mathcal{E}_{\mathit{batch}}}} \sum_{\bm{e} \in \mathcal{E}_{\mathit{batch}}} \lambda_{\mathit{diff}} \ell_{\mathit{diff}} (\bm{e}) + \lambda_{\mathit{grad}} \ell_{\mathit{grad}} (\bm{e}) \ .
    \label{eq:ren_total_loss}
\end{equation}

\noindent
The \textit{threshold-normalized difference loss} $\ell_{\mathit{diff}}$ with weight $\lambda_{\mathit{diff}}$ acts as the main reconstruction loss. It enforces the \textit{mean contrast threshold} $\bar{C} = \frac{1}{2} (C_{-1} + C_{+1})$ normalized squared consistency between the predicted log-radiance difference $\Delta \log \hat{L} \defeq \log \hat{L} (\bm{u}, t_{\mathit{curr}}) - \log \hat{L} (\bm{u}, t_{\mathit{ref}})$, given by NeRF renders, and the observed log-radiance difference $\Delta \log L = p C_p$ from an event (\cref{eq:egm}), as follows:
\begin{equation}
	\ell_{\mathit{diff}} (\bm{e}) = \left( \frac{\Delta \log \hat{L} - p C_p}{\bar{C}} \right)^2 \ .
    \label{eq:l_diff}
\end{equation}

\noindent
The \textit{target-normalized gradient loss} $\ell_{\mathit{grad}}$ with weight $\lambda_{\mathit{grad}}$ acts as a smoothness constraint for regularization of textureless regions. It represents the \textit{Absolute Percentage Error} $\operatorname{APE} \left(\hat{y}, y \right) = \abs*{\nicefrac{\hat{y} - y}{y}}$ between the predicted log-radiance gradient $\frac{\partial}{\partial t} \log \hat{L} (\bm{u}, t)$, computed using auto-differentiation, and the finite difference approximation of the target log-radiance gradient $\frac{\partial}{\partial t} \log L (\bm{u}, t) \approx \frac{p C_p}{t_{\mathit{curr}} - t_{\mathit{ref}}}$, at a timestamp $t_{\mathit{sam}}$ sampled between $t_{\mathit{ref}}$ and $t_{\mathit{curr}}$, as follows:

\begin{equation}
	\ell_{\mathit{grad}} (\bm{e}) = \operatorname{APE} \left(\frac{\partial}{\partial t} \log \hat{L} (\bm{u}, t_{\mathit{sam}}) \fraccomma \ \frac{p C_p}{t_{\mathit{curr}} - t_{\mathit{ref}}} \right) \ .
    \label{eq:l_grad}
\end{equation}

\bfPar{Gamma Correction.}
Since event cameras mainly provide observations of \textit{changes} in log-radiance, not \emph{absolute} log-radiance, the predicted log-radiance $\log \hat{\bm{L}}$ from the reconstructed NeRF is only accurate up to an offset per color channel. There will be an additional channel-consistent scale ambiguity, when only the \textit{Contrast Threshold Ratio} $\nicefrac{C_{+1}}{C_{-1}}$ is known during reconstruction. Nonetheless, these ambiguities can be resolved post-reconstruction, given a set of reference images. Specifically, \textit{ordinary least squares} can be used to perform an \textit{affine} correction on $\log \hat{\bm{L}}$, or equivalently a \textit{gamma} correction on $\hat{\bm{L}}$, as follows:
\begin{equation}
    \log \hat{\bm{L}}_\mathit{corr} = a \odot \log \hat{\bm{L}} + \bm{b} \ ,
    \label{eq:gamma_corr}
\end{equation}
where $a$ and $\bm{b}$ are the correction parameters.

\section{Our Method}
\label{sec:method}

We first introduce the physically-accurate pixel bandwidth model proposed to model the motion blur of events (\cref{sec:method:pbm}), which extends the event generation model of Robust \textit{e}-NeRF (\cref{sec:prelims}) . Subsequently, we detail how to synthesize motion-blurred (effective) log-radiance incident at a pixel (\cref{sec:method:synthesis}), thus event motion blur, for optimization of a blur-minimal NeRF from motion-blurred events (\cref{sec:method:training}). Lastly, we present an enhanced variant of gamma correction (\cref{sec:prelims}) that takes the pixel bandwidth model into consideration (\cref{sec:method:t_gamma_corr}).

\subsection{Pixel Bandwidth Model}
\label{sec:method:pbm}

\begin{figure}[t]
    \centering

    \begin{minipage}[t]{0.32\textwidth}
    \centering
        \includegraphics[width=1\linewidth,trim={0.4cm 0.3cm 0.4cm 0.2cm},clip]{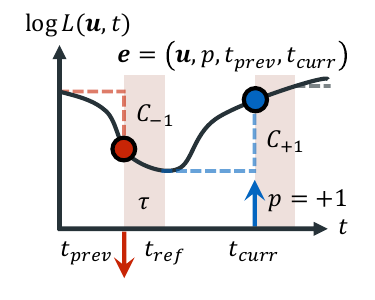}
        
        \caption{Robust \textit{e}-NeRF event generation model \cite{low2023_robust-e-nerf} }
        \label{fig:egm}
    \end{minipage} \quad
    \begin{minipage}[t]{0.64\textwidth}
    \centering
        \includegraphics[width=1\linewidth]{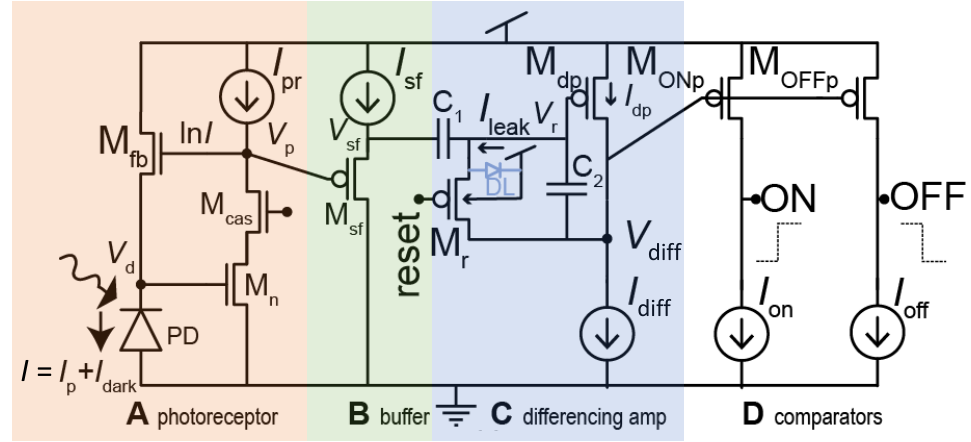}
        
        \caption{Core analog circuit of a typical event sensor pixel (Adapted from \cite{nozaki2017_temperature})}
        \label{fig:pixel_circuit}
    \end{minipage}
\end{figure}

\begin{figure}[t]
    \centering
    \includegraphics[width=1.0\linewidth]{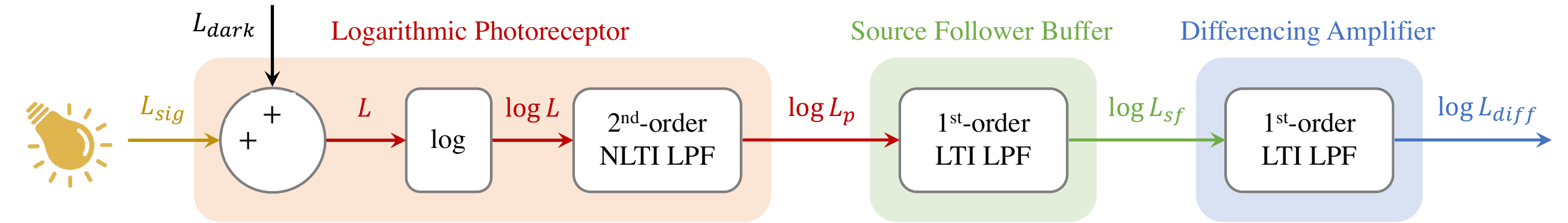}
    
    \caption{Overview of the proposed pixel bandwidth model}
    \label{fig:pbm}
\end{figure}

Event motion blur is attributed to the limited bandwidth of the sensor pixel analog circuit, which also bounds the minimum event detection latency, maximum frequency of change detectable and hence maximum event generation rate.

\bfPar{Event Pixel Circuit.}
\cref{fig:pixel_circuit} shows the core analog circuit of a typical event sensor pixel (in particular, that of the DVS128 \cite{lichtsteiner2008_128}), which consists of 4 stages.

The \textit{Logarithmic Photoreceptor} (Stage A) contains a \textit{Photodiode} (PD) that transduces radiance (more accurately, irradiance) incident at the pixel to \textit{Signal Photocurrent} $I_p$ proportionally, and an active feedback loop that outputs a voltage $V_p$ proportional to the logarithm of the \textit{Photocurrent} $I = I_p + I_\mathit{dark}$ \cite{lichtsteiner2008_128,graca2023_shining,nozaki2017_temperature}. A small \textit{Dark Current} $I_\mathit{dark}$ flows through the photodiode, even in the dark.

Next, $V_p$ is buffered with a \textit{Source Follower} (Stage B) to isolate the sensitive photoreceptor from rapid transients in subsequent stages \cite{lichtsteiner2008_128, nozaki2017_temperature}. The \textit{Differencing Amplifier} (Stage C) then amplifies the change in source follower buffer output $V_\mathit{sf}$ from the reset/reference voltage level, \& outputs a voltage $V_\mathit{diff}$ to be compared with both ON \& OFF thresholds for event detection (Stage D) \cite{lichtsteiner2008_128, graca2023_shining, nozaki2017_temperature}. When $V_\mathit{diff}$ exceeds either thresholds, an event is generated and the differencing amplifier is held in reset for a duration of the refractory period $\tau$ \cite{lichtsteiner2008_128, graca2023_shining, nozaki2017_temperature}.

\bfPar{Model.}
The design of the pixel analog circuit entails that event cameras in fact respond to changes in \textit{effective log-radiance} $\log L = \log \left( L_\mathit{sig} + L_\mathit{dark} \right)$ instead, where $L_\mathit{sig}$ is the actual incident radiance signal and $L_\mathit{dark}$ is the \textit{black level}. Similar to standard image sensors, the black level is defined as the dark current-equivalent incident radiance, which is exponentially sensitive to temperature \cite{nozaki2017_temperature} and effectively limits the dynamic range of the sensor \cite{graca2023_shining}.

More precisely, as the pixel bandwidth is mainly limited by the first 3 stages of the analog circuit \cite{lichtsteiner2006_aer,mcreynolds2022_experimental, delbruck1995_phototransduction,delbruck1993_thesis,graca2021_unraveling, graca2023_optimal_biasing}, event cameras effectively measure changes in the \textit{low-pass-filtered/motion-blurred effective log-radiance} $\log L_\mathit{blur}$. This explains the discussed motion blur of events and event sensor dynamic performance limit.

We accurately model the band-limiting behavior of the pixel with a unity-gain 4\textsuperscript{th}-order \textit{Non-Linear Time-Invariant} (NLTI) \textit{Low-Pass Filter} (LPF) in state-space form, with input $u = \log L$, state
$\bm{x} = \begin{bsmallmatrix}
    \nicefrac{\partial \log L_\mathit{p}}{\partial t} &
    \log L_p &
    \log L_\mathit{sf} &
    \log L_\mathit{diff}
 \end{bsmallmatrix}^\top$ 
and output 
$\bm{y} = \begin{bsmallmatrix}
    \log L_\mathit{sf} &
    \log L_\mathit{diff}
 \end{bsmallmatrix}^\top$,
as follows:

\begin{equation}
    \begin{aligned}
        \dot{\bm{x}} ( t ) &= A \left( u(t) \right) \ \bm{x} ( t ) + B \left( u(t) \right) \ u(t) \\
        \bm{y} (t) &= C \ \bm{x} ( t )
    \end{aligned} \ ,
    \label{eq:pbm}
\end{equation}

\begin{flalign*}
    \text{where }
    A ( u ) &=
    \begin{bmatrix}
        -2 \zeta ( u ) \omega_n ( u ) & -\omega_n^2 ( u ) & 0 & 0 \\
        1 & 0 & 0 & 0 \\
        0 & \omega_\mathit{c, sf} & -\omega_\mathit{c, sf} & 0 \\
        0 & 0 & \omega_\mathit{c, diff} & -\omega_\mathit{c, diff}
    \end{bmatrix}
    , \ 
    B ( u ) =
    \begin{bmatrix}
        \omega_n^2 ( u ) \\
        0 \\
        0 \\
        0
    \end{bmatrix}
    , &\\
    C &=
    \begin{bmatrix}
        0 & 0 & 1 & 0 \\
        0 & 0 & 0 & 1
    \end{bmatrix} . &
\end{flalign*}
Specifically, the pixel bandwidth model is formed by a cascade of:

\begin{enumerate}
    \item A unity-gain 2\textsuperscript{nd}-order NLTI LPF, with input $\log L$, state
        $\begin{bsmallmatrix}
            \nicefrac{\partial \log L_p}{\partial t} &
            \log L_p
        \end{bsmallmatrix}^\top$ 
        and output $\log L_\mathit{p}$, that models the transient response of the logarithmic photoreceptor \cite{lichtsteiner2008_128,lichtsteiner2006_aer,mcreynolds2022_experimental, delbruck1995_phototransduction,delbruck1993_thesis,graca2021_unraveling, graca2023_optimal_biasing}. Similar to its \textit{Linear Time Invariant} (LTI) counterpart, this 2\textsuperscript{nd}-order filter is characterized by its \textit{Damping Ratio} $\zeta$ and \textit{Natural Angular Frequency} $\omega_n$. However, they are not constants, but complex non-linear functions of its input \cite{delbruck1993_thesis,delbruck1995_phototransduction,lichtsteiner2008_128,lichtsteiner2006_aer} (more details in the supplement). The bandwidth of this filter is mostly proportional to the exponential of its input $\exp u = L$, which explains the susceptibility to motion blur under low-light. However, black level $L_\mathit{dark}$ limits the minimum pixel bandwidth.
        \item A unity-gain 1\textsuperscript{st}-order LTI LPF, with input $\log L_p$ and state/output $\log L_\mathit{sf}$, that models the transient response of the source follower buffer \cite{lichtsteiner2006_aer,mcreynolds2022_experimental,graca2023_optimal_biasing}. It is characterized by its constant bandwidth/\textit{Cutoff Angular Frequency} $\omega_\mathit{c, sf}$, that is proportional to the \textit{source follower buffer bias current} $I_\mathit{sf}$ (\cref{fig:pixel_circuit}) \cite{lichtsteiner2006_aer}.
        \item Another unity-gain 1\textsuperscript{st}-order LTI LPF, with input $\log L_\mathit{sf}$, state/output $\log L_\mathit{diff}$ and cutoff angular frequency $\omega_\mathit{c, diff} > \omega_\mathit{c, sf}$ \cite{lichtsteiner2006_aer}, that models the transient response of the differencing amplifier \cite{lichtsteiner2006_aer,mcreynolds2022_experimental}.

\end{enumerate}
as illustrated in \cref{fig:pbm}.

We also model the steady-state behavior of the differencing amplifier reset mechanism as a reset of the amplifier LPF state/output $\log L_\mathit{diff}$ to its input $\log L_\mathit{sf}$, at the end of the refractory period (\ie reference timestamp $t_\mathit{ref}$). These 2 models allow the motion-blurred effective log-radiance $\log L_\mathit{blur}$ to be derived as:
\begin{equation}
    \log L_\mathit{blur}(t) = \log L_\mathit{diff}(t) + \log L_\mathit{delta}(t_\mathit{ref}) \ e^{- \omega_\mathit{c, diff} \left( t - t_\mathit{ref} \right)} \ , \ t \geq t_\mathit{ref} \ ,
    \label{eq:logL_blur}
\end{equation}
where $\log L_\mathit{delta} = \log L_\mathit{sf} - \log L_\mathit{diff}$.

\subsection{Synthesis of Motion-Blurred Effective Log-Radiance}
\label{sec:method:synthesis}

\begin{figure}[t]
    \centering
    \includegraphics[width=1.0\linewidth]{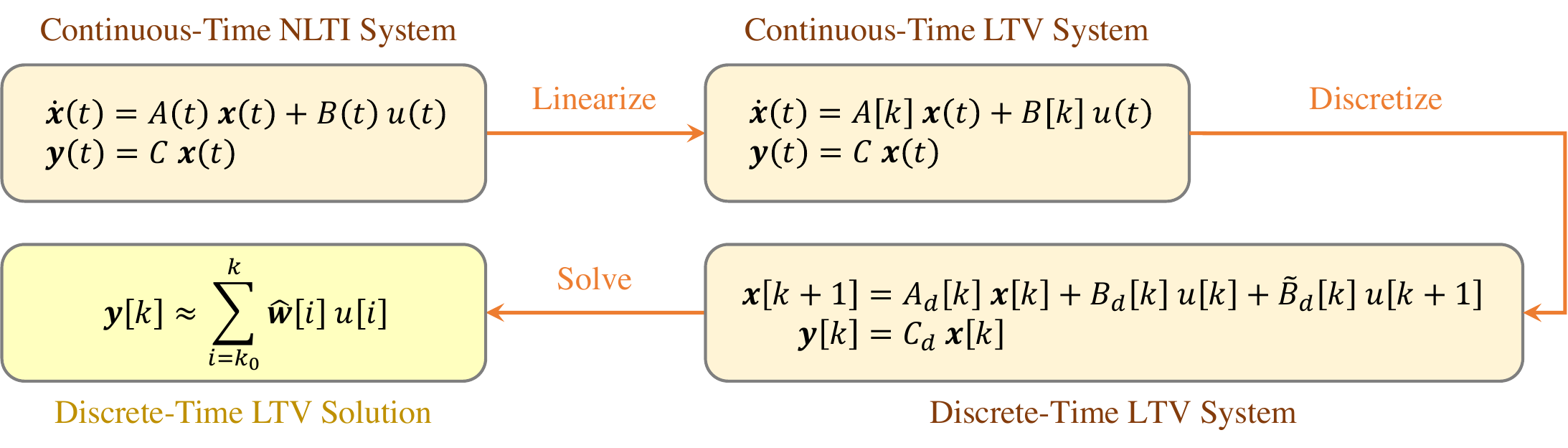}
    
    \caption{Overview of deriving the numerical solution of the pixel bandwidth model}
    \label{fig:synthesis}
\end{figure}

The pixel bandwidth model proposed in \cref{sec:method:pbm} provides a means to accurately synthesize motion-blurred effective log-radiance $\log L_\mathit{blur}$, thus simulate event motion blur, given the pixel-incident log-radiance signal $\log L_\mathit{sig}$. However, the continuous-time and non-linear nature of the model (\cref{eq:pbm}) prohibits its direct computational implementation, for use in a simulator or for NeRF reconstruction using an Analysis-by-Synthesis framework (\cref{sec:method:training}). A discrete-time counterpart of the model that operates on discrete-time input samples is necessary (\cref{fig:synthesis}).

\bfPar{Discrete-Time Model.}
Assume the discrete-time sequence of inputs $u [k] = \log L [k]$ is sampled at timestamps $t_k$, where the time intervals between successive samples $\delta t_k = t_{k+1} - t_k$ may possibly be irregular. We derive a discrete-time model from the continuous-time, non-linear pixel bandwidth model by first linearizing \cite{franklin1998_digital_control} the 4\textsuperscript{th}-order NLTI LPF (\cref{eq:pbm}) at different steady-state operating points 
    $\left( \bar{\bm{x}}  [k], \bar{u} [k] \right)
     = \left( 
        \begin{bsmallmatrix}
            0 &
            u [k+1] &
            u [k+1] &
            u [k+1]
        \end{bsmallmatrix}^\top
     , u [k+1] \right)$, 
    for each time interval $\left( t_k, t_{k+1} \right]$. Then, we discretize the linearized model assuming \textit{First-Order Hold} (FOH) \cite{franklin1998_digital_control} (\ie piecewise-linear) inputs $u$. This yields a discrete-time 4\textsuperscript{th}-order \textit{Linear Time-Varying} (LTV) LPF, in non-standard state-space form, as follows:
\begin{equation}
    \begin{aligned}
        \bm{x} [k+1] &= A_d [k] \ \bm{x} [k] + B_d [k] \ u [k] + \widetilde{B}_d [k] \ u [k+1] \\
        \bm{y} [k] &= C_d \ \bm{x} [k]
    \end{aligned} \ ,
    \label{eq:discretized_pbm}
\end{equation}
where $\ A_d [k] = \Phi[k], \ B_d [k] = \Gamma_1[k] - \Gamma_2[k], \ \ \widetilde{B}_d [k] = \Gamma_2[k], \ C_d = C \text{ and:}$
\begin{equation*}
    \begin{bmatrix}
        \Phi [k] & \Gamma_1 [k] & \Gamma_2 [k] \\
        0 & I & I \\
        0 & 0 & I
    \end{bmatrix} =
    \exp{\left(
        \begin{bmatrix}
            A \left( u[k+1] \right) \delta t_k & B \left( u[k+1] \right) \delta t_k & 0 \\
            0 & 0 & I \\
            0 & 0 & 0
        \end{bmatrix}
    \right)} \ .
\end{equation*}

\bfPar{Numerical Solution.}
The discrete-time model presented can be directly integrated in an existing event simulator, \eg ESIM \cite{rebecq2018_esim} \& its improved variant introduced in Robust \textit{e}-NeRF, to synthesize $\log L_\mathit{blur}$ \& thus simulate event motion blur (more details in the supplement), assuming some appropriate initial state $\bm{x} [k_0]$, \eg the steady-state on the initial input $u [k_0]$. However, this cannot be done when the appropriate $\bm{x} [k_0]$ is not well defined for an arbitrary $\log L_\mathit{blur} [k]$ of interest, \eg during NeRF reconstruction. We tackle this issue with the (numerical) solution to the transient response of the discrete-time model below:
\begin{equation}
    \bm{y} [k] = C_d \left[ \varphi(k_0, k) \ \bm{x} [k_0] + \sum_{i = k_0}^{k-1} {  \varphi(i+1, k) \left( B_d[i] \ u[i] + \widetilde{B}_d[i] \ u[i+1] \right) } \right] \ ,
    \label{eq:pbm_sol}
\end{equation}
where the \textit{state transition matrix} $\varphi(m, n) = \prod_{j=1}^{n-m} A_d [n-j]$.

As the linearized, thus discretized, model is \textit{asymptotically stable}, the magnitude of eigenvalues of $A_d [k]$, for all $k$, must be smaller than 1. This entails that $\lim_{k-k_0 \rightarrow \infty}{\varphi(k_0, k)} = 0$. Thus, for a sufficiently long numerical integration time interval $\left( t_{k_0}, t_k \right]$, $\bm{y} [k]$ can be approximated as the \textit{zero-state response} of the model, which is just a weighted sum of past \& present inputs $u$ between $k_0$ \& $k$:
\begin{equation}
    \bm{y} [k] \approx \sum_{i = k_0}^{k} \bm{w} [i] \ u[i] \approx \sum_{i = k_0}^{k} \hat{\bm{w}} [i] \ u[i] \ ,
    \label{eq:approx_pbm_sol}
\end{equation}
\begin{align*}
    \text{where }
    \bm{w} [i] &= 
    \begin{cases}
        C_d \varphi(k_0 + 1, k) B_d [k_0] & ,\text{if } i = k_0 \\
        C_d \left( \varphi(i+1, k) B_d[i] + \varphi(i, k) \widetilde{B}_d[i-1] \right) & ,\text{if } i = k_0 + 1, \ \ldots, \ k-1 \\
        C_d \widetilde{B}_d[k-1] & ,\text{if } i = k
    \end{cases} \\
    % \hat{\bm{w}} [i] &= \frac{\bm{w} [i]}{\sum_{j = k_0}^{k} \bm{w} [j]} \ .
    \hat{\bm{w}} [i] &= \bm{w} [i] \oslash \sum_{j = k_0}^{k} \bm{w} [j] \ ,
\end{align*}
and $\oslash$ denotes \textit{Hadamard}/element-wise division. It can be shown that $\lim_{k-k_0 \rightarrow \infty}$ ${\sum_{i = k_0}^{k} \bm{w} [i] = \bm{1}}$. Thus, we use the \textit{sum-normalized weights} $\hat{\bm{w}} [i]$ in practice, as they are corrected for the bias due to a finite integration interval $\left( t_{k_0}, t_k \right]$. In general, these weights tend to be larger for larger input (\ie higher effective log-radiance) samples with timestamps closer to the output timestamp $t_k$.

\bfPar{Importance Sampling.}
Often times, we are interested in computing $\log L_\mathit{blur}$ at some desired timestamp $t_k$, given only a function for sampling inputs $u = \log L$ and a fixed sampling budget. To this end, we propose to infer the optimal input sample timestamps, represented by the random variable $T_i \in \left( -\infty, t_k \right]$, by sampling them from the following \textit{transformed exponential distribution}:
\begin{equation}
    T_i \sim \operatorname{Exp}{\left( t_k - t_i ; \ \omega_\mathit{c, dom, min} \right)} = \omega_\mathit{c, dom, min} \ e^{-\omega_\mathit{c, dom, min} \left( t_k - t_i \right)} \ , 
    \label{eq:tf_exp_proposal}
\end{equation}
where $\omega_\mathit{c, dom, min}$ is the minimum possible dominant cutoff angular frequency of the pixel, achieved under extreme low-light when $L = L_\mathit{dark}$.

The suggested proposal distribution coarsely approximates the distribution represented by $\hat{\bm{w}} [i]$ over the interval $\left( t_{k_0}, t_k \right]$, as it corresponds to the weight function, derived from the zero-state response of the continuous-time dominant pole-approximated model under extreme low-light. Thus, it generally concentrate samples at relevant parts of the input (\ie with large weight), achieving a similar goal as \textit{importance sampling}. It works best under low-light, when event cameras are most susceptible to motion blur. More details are available in the supplement.

\subsection{Training}
\label{sec:method:training}

We employ the same training procedure used in Robust \textit{e}-NeRF (\cref{sec:prelims}) to optimize a blur-minimal NeRF from motion-blurred events in an Analysis-by-Synthesis manner, with a few exceptions. Specifically, we apply our training losses on the predicted motion-blurred effective log-radiance $\log \hat{L}_\mathit{blur}$, which was synthesized from NeRF renders $\hat{L}$ using the proposed numerical solution to the pixel bandwidth model (\cref{eq:approx_pbm_sol,eq:logL_blur}) with importance sampling (\cref{eq:tf_exp_proposal}).

Moreover, we adopt a \textit{Huber}-norm ($\delta = 1$) variant of the threshold-normalized difference loss $\ell_{\mathit{diff}}$  (\cref{eq:l_diff}), which is less sensitive to outliers, with weight $\lambda_{\mathit{diff}}$. We also propose the \textit{threshold-normalized total variation loss} $\ell_{\mathit{tv}}$, as a replacement for the target-normalized gradient loss $\ell_{\mathit{grad}}$ (\cref{eq:l_grad}), with weight $\lambda_{\mathit{tv}}$.

\bfPar{Fundamental Limitations.} When the black level $L_\mathit{dark}$ is unknown, the incident radiance signal $L_\mathit{sig}$ and $L_\mathit{dark}$ cannot be unambiguously disentangled from just the observation of effective radiance $L = L_\mathit{sig} + L_\mathit{dark}$ given by events. Thus, we can only reconstruct a NeRF with volume renders $\hat{\bm{L}}$ that represent predicted effective radiance, not predicted incident radiance signal as suggested by Robust \textit{e}-NeRF. This is enabled by the assumption that the temperature of the event camera remains effectively constant over the entire duration of the given event steam, so that the dark current \& thus black level remains effectively stationary.

Furthermore, as the pixel bandwidth depends on the \textit{absolute} effective radiance $L$, the predicted effective radiance $\hat{\bm{L}}$ is theoretically gamma-accurate (\ie gamma correction of $\hat{\bm{L}}$ is unnecessary), assuming known pixel bandwidth model parameters, contrary to what is suggested by Robust \textit{e}-NeRF. However, in practice, $\hat{\bm{L}}$ is generally only gamma-accurate if the $L$ associated to the events has a significant impact on the pixel transient response. In other words, the gamma-accuracy of $\hat{\bm{L}}$ greatly depends on the severity of event motion blur, hence camera speed, scene illumination, scene texture complexity, camera used and its settings.

\bfPar{Threshold-Normalized Total Variation Loss.} This loss penalizes the \textit{mean contrast threshold} $\bar{C} = \frac{1}{2} (C_{-1} + C_{+1})$ normalized \textit{total variation} of the predicted motion-blurred effective log-radiance $\log \hat{L}_\mathit{blur}$, on a subinterval $\left( t_\mathit{start}, t_\mathit{end} \right]$ sampled between the interval $\left( t_\mathit{ref}, t_\mathit{curr} \right]$ given by an event, as follows:
\begin{equation}
	\ell_{\mathit{tv}} (\bm{e}) = \abs*{ \frac{\delta \log \hat{L}_\mathit{blur}}{\bar{C}} } \ ,
    \label{eq:l_tv}
\end{equation}
where $\delta \log \hat{L}_\mathit{blur} \defeq \log \hat{L}_\mathit{blur} (\bm{u}, t_{\mathit{end}}) - \log \hat{L}_\mathit{blur} (\bm{u}, t_{\mathit{start}})$.

Similar to $\ell_{\mathit{grad}}$  (\cref{eq:l_grad}), this loss acts as a smoothness constraint for regularization of textureless regions in the scene. However, it imposes a stronger bias to enforce the uniformity of $\log \hat{L}_\mathit{blur}$ between event intervals, which greatly helps with reconstructing large uniform patches. It can also effectively generalize to arbitrary threshold values due to the normalization, similar to $\ell_{\mathit{diff}}$  (\cref{eq:l_diff}).

\bfPar{Joint Optimization of Pixel Bandwidth Model Parameters.}
Our method does not strictly rely on the \textit{pixel bandwidth model parameters} $\Omega$ to be known as \textit{a priori}, as it generally supports their joint optimization with \textit{NeRF parameters} $\Theta$. However, their prior knowledge generally facilitates a more accurate reconstruction. The parameters $\Omega$, which include the parameters of non-linear damping ratio function $\zeta$ \& natural angular frequency function $\omega_n$, and angular cutoff frequencies $\omega_\mathit{c, sf}$ and $\omega_\mathit{c, diff}$ (\cref{eq:pbm}), depend on the pixel circuit design, semiconductor manufacturing process \& user-defined event camera bias settings.

\subsection{Translated-Gamma Correction}
\label{sec:method:t_gamma_corr}

To eliminate the unknown black level offset and resolve potential gamma-inacc\-uracies in the predicted effective radiance $\hat{\bm{L}}$, we propose a \textit{Translated-Gamma Correction} on $\hat{\bm{L}}$ post-reconstruction, using a set of reference images, as follows:
\begin{equation}
    \hat{\bm{L}}_\mathit{sig, corr} = \bm{b} \odot \hat{\bm{L}}^a - \bm{c} \ ,
    \label{eq:translated_gamma_corr}
\end{equation}
where $a$, $\bm{b}$ and $\bm{c}$ are the correction parameters, via \textit{Levenberg-Marquardt} non-linear least squares optimization. The translation/offset correction is done independently per color channel to account for channel-varying spectral sensitivities.

\section{Experiments}
\label{sec:exp}

We conduct a series of \textit{novel view synthesis} experiments, both on synthetic (\cref{sec:exp:synthetic}) and real event sequences (\cref{sec:exp:real}), to verify that our method, Deblur \textit{e}-NeRF, can indeed directly and effectively reconstruct blur-minimal NeRFs from motion-blurred events, generated under high-speed or low-light conditions, using a physically-accurate pixel bandwidth model.

\bfPar{Metrics.}
We adopt the commonly used PSNR, SSIM \cite{zhou2004_ssim} and AlexNet-based LPIPS \cite{zhang2018_lpips} to evaluate the similarity between the target and translated-gamma-corrected (\cref{sec:method:t_gamma_corr}) synthesized novel views, for all methods in each experiment.

\bfPar{Baselines.}
We benchmark our method, Deblur \textit{e}-NeRF, against the state-of-the-art, Robust \textit{e}-NeRF \cite{low2023_robust-e-nerf}, and a naïve baseline of E2VID \cite{rebecq2021_e2vid} (a seminal events-to-video reconstruction method) $+$ NeRF \cite{mildenhall2020_nerf} (as well as 2 other \textit{image} blur-aware baselines in the supplement). The setup of an \textit{event} motion blur-aware baseline is hindered by the lack of relevant works. The implementation of all methods employ a common NeRF backbone \cite{li2023_nerfacc} to allow for a fair comparison.

\bfPar{Datasets.}
We perform the synthetic experiments on the default set of sequences released by Robust \textit{e}-NeRF, and a novel set similarly simulated on the ``Realistic Synthetic $360^\circ$'' scenes \cite{mildenhall2020_nerf}. However, our sequences involve event motion blur due to fast camera motion and/or poor scene illumination. Such sequences were simulated with our proposed event simulator, which incorporates the pixel bandwidth model presented in \cref{sec:method:pbm,sec:method:synthesis} in the improved ESIM \cite{rebecq2018_esim} event simulator introduced in Robust \textit{e}-NeRF. Similar to the sequences in the Robust e-NeRF synthetic event dataset, the events are generated from a virtual event camera moving in a hemi-/spherical spiral motion about the object at the origin.

On the other hand, we conduct the real experiments on the \texttt{08\_peanuts\_runn}\- \texttt{ing} and \texttt{11\_all\_characters} sequences of the EDS dataset \cite{hidalgo2022_eds}, which are generally $360^\circ$ captures of a table of objects in an office room under moderate lighting. These sequences were chosen as they involve some high-speed camera motion.

\subsection{Synthetic Experiments}
\label{sec:exp:synthetic}

\begin{figure}[t]
    \centering
    \includegraphics[width=0.56\linewidth]{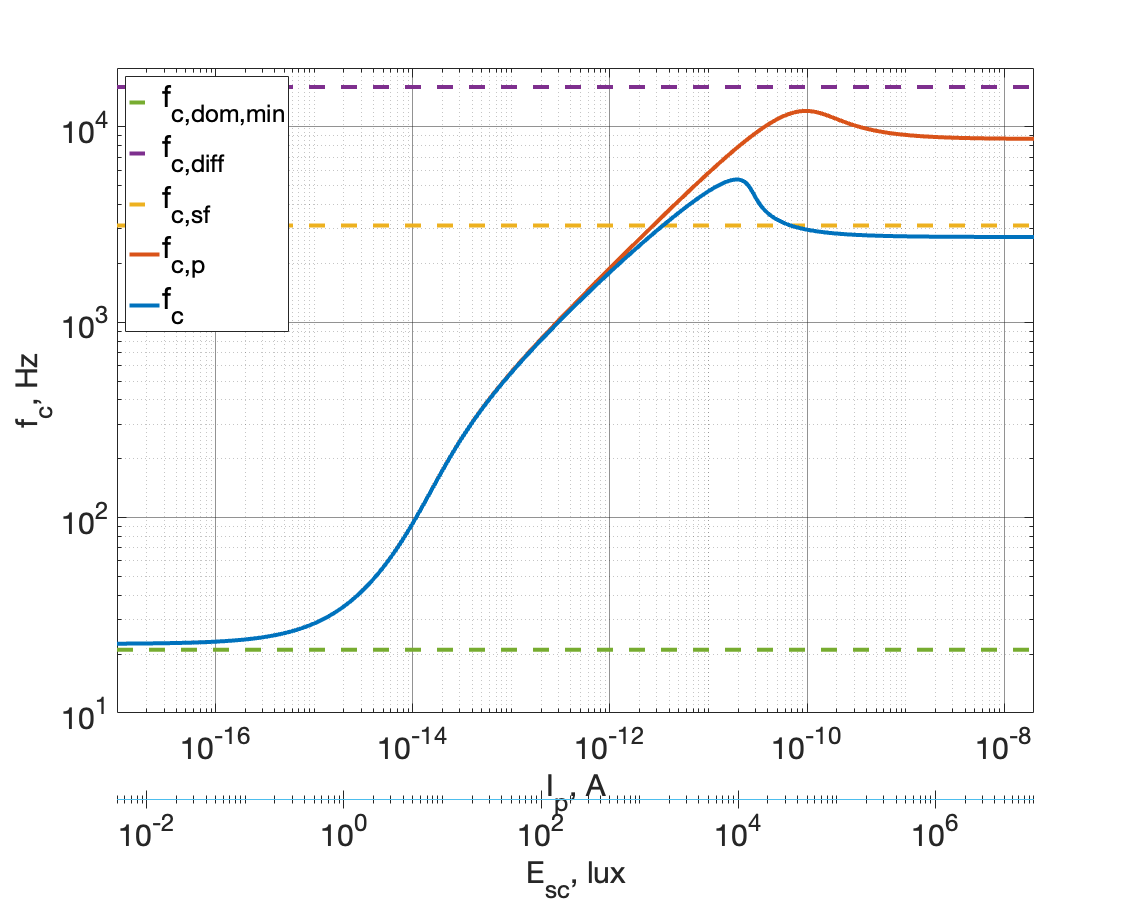}
    
    \caption{Pixel bandwidth of DVS128 \cite{lichtsteiner2008_128} with nominal biases}
    \label{fig:dvs128_pix_bw}
\end{figure}

The synthetic experiments serve as the main benchmark to assess all methods, as they enable controlled tests under diverse realistic conditions with precise ground truth, a task that would otherwise be infeasible using real sequences.

As with the default sequences in Robust \textit{e}-NeRF \cite{low2023_robust-e-nerf}, we simulate ours with symmetric contrast thresholds of $0.25$ (\ie $C_{-1} = C_{+1} = 0.25$), zero pixel-to-pixel threshold standard deviation and refractory period (\ie $\sigma_{C_p} = 0, \tau = 0$), and provide camera poses at $1 \ \mathit{kHz}$. Moreover, the pixel bandwidth model parameters that we adopt correspond to that of the DVS128 \cite{lichtsteiner2008_128} event camera with nominal biases (provided in jAER \cite{delbruck2008_jaer}). \cref{fig:dvs128_pix_bw} illustrates the light-dependent behaviour of its pixel bandwidth, which is mostly proportional to the \textit{scene illuminance} $E_\mathit{sc}$, thus radiance incident at a pixel $L_\mathit{sig}$. Unless otherwise stated, the virtual event camera revolves the object 4 times with uniform 2 revolution per second speed about the object vertical axis. All sequences are also simulated under a \textit{scene illuminance} $E_\mathit{sc}$ of $1 \ 000 lux$ by default, which corresponds to standard office lighting \cite{delbruck1995_phototransduction,delbruck1993_thesis}. Under such a condition, the pixel bandwidth spans around $50$--$2500 \mathit{Hz}$, depending on the incident radiance. Due to limited resources, our method is only trained with $\nicefrac{1}{8} \times$ the batch size of our baselines, by default.

\begin{figure}[t!]
    \centering   
    \includegraphics[width=1\linewidth]{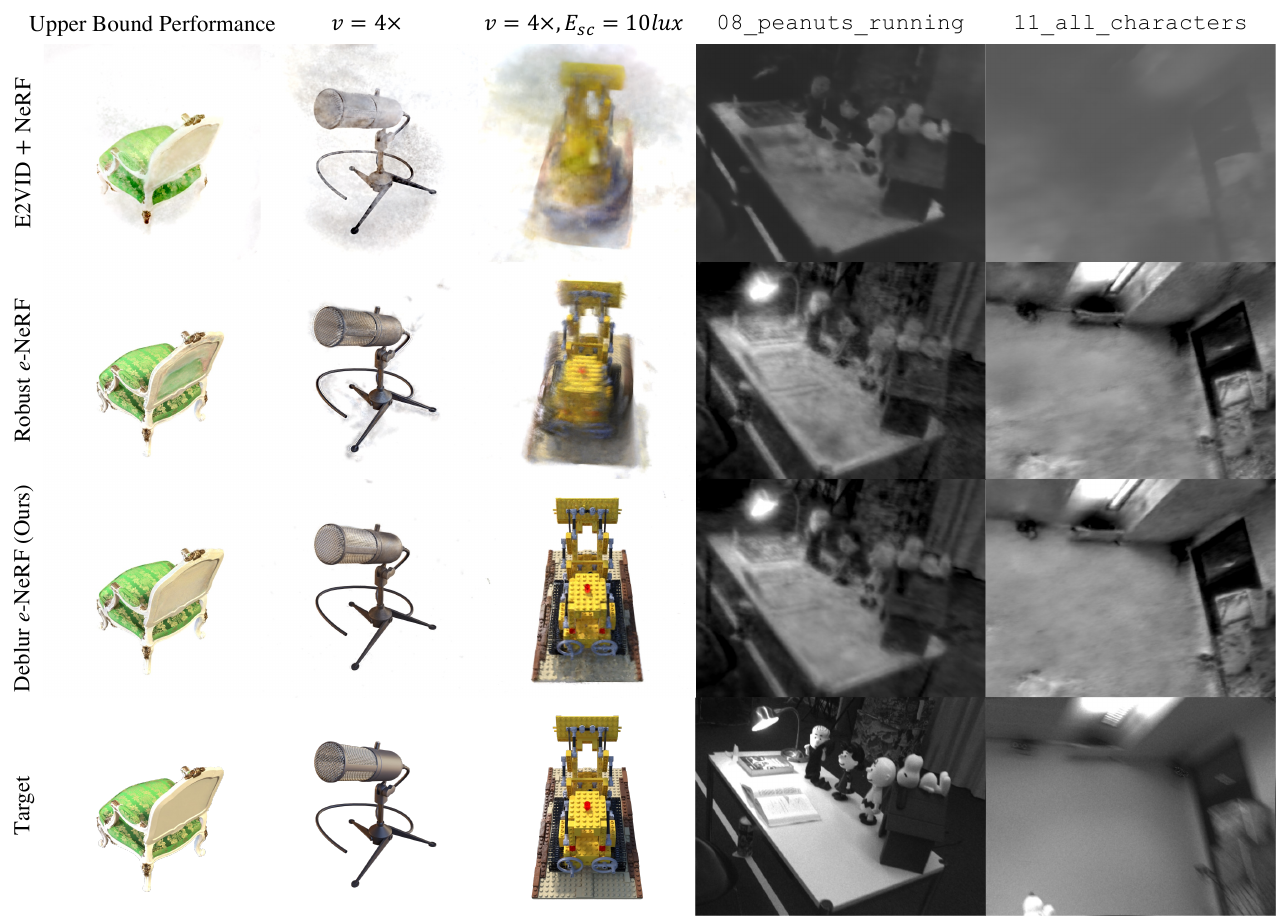}

    \caption{Qualitative results of synthetic \& real experiments, w/o jointly optimized contrast thresholds, refractory period and pixel bandwidth model parameters, if applicable}
    \label{fig:qualitative}
\end{figure}

\bfPar{Upper Bound Performance.}
To quantify the upper bound performance, we evaluate all methods on motion blur-free event sequences, which is effectively given by a pixel with infinite bandwidth. For this purpose, we remove the pixel bandwidth model from our method and train it with the same batch size as the baselines. The main difference between Robust \textit{e}-NeRF and our method, under such a setting, is the replacement of the target-normalized gradient loss $\ell_{\mathit{grad}}$ with the threshold-normalized total variation loss $\ell_{\mathit{tv}}$. Thus, the quantitative results reported in \cref{tab:upper_bound} verifies the effectiveness of our proposed $\ell_{\mathit{tv}}$ over $\ell_{\mathit{grad}}$. Moreover, the qualitative results shown in \cref{fig:qualitative}, particularly at the back of the chair, clearly shows the strength of $\ell_{\mathit{tv}}$ in regularizing large textureless patches.

\begin{table}[t!]
\fontsize{5.3}{6.276}
\selectfont

\begin{minipage}{0.38\linewidth}
    \centering
    \caption{Upper bound performance without event motion blur}
    \label{tab:upper_bound}

    \begin{tabular}{@{}llccc@{}}
    \toprule
    \multicolumn{1}{c}{Method} &  & PSNR $\uparrow$ & SSIM $\uparrow$ & LPIPS $\downarrow$ \\ \midrule
    E2VID $+$ NeRF &  & 19.49 & 0.847 & 0.268 \\
    Robust \textit{e}-NeRF &  & 28.48 & 0.944 & 0.054 \\
    \rowcolor[HTML]{F3F3F3} 
    Deblur \textit{e}-NeRF & & \textbf{29.43} & \textbf{0.953} & \textbf{0.043} \\ \bottomrule
    \end{tabular}
\end{minipage}\hfill
\begin{minipage}{0.6\linewidth}
    \centering
    \caption{Quantitative results of the real exps.}
    \label{tab:real}

    \begin{tabular}{@{}lccclccc@{}}
    \toprule
    \multicolumn{1}{c}{} & \multicolumn{3}{c}{\texttt{08\_peanuts\_running}} &  & \multicolumn{3}{c}{\texttt{11\_all\_characters}} \\ \cmidrule(lr){2-4} \cmidrule(l){6-8} 
    \multicolumn{1}{c}{\multirow{-2}{*}{Method}} & PSNR $\uparrow$ & SSIM $\uparrow$ & LPIPS $\downarrow$ &  & PSNR $\uparrow$ & SSIM $\uparrow$ & LPIPS $\downarrow$ \\ \midrule
    E2VID $+$ NeRF & 14.85 & 0.690 & 0.595 &  & 13.12 & 0.695 & 0.627 \\
    Robust \textit{e}-NeRF & 18.00 & 0.677 & 0.507 &  & 15.91 & 0.677 & 0.552 \\
    \rowcolor[HTML]{F3F3F3} 
    Deblur \textit{e}-NeRF & \textbf{18.27} & \textbf{0.695} & \textbf{0.503} &  & \textbf{16.53} & \textbf{0.710} & \textbf{0.511} \\ \bottomrule
    \end{tabular}
\end{minipage}

\end{table}

\begin{table*}[t!]
\fontsize{7.5}{9}
\selectfont

\centering
\caption{Effect of camera speed. $^\dag$Trained with $\nicefrac{1}{8} \times$ the batch size of baselines.}
\label{tab:speed}

\begin{tabular}{@{}lccclccclccc@{}}
\toprule
\multicolumn{1}{c}{} & \multicolumn{3}{c}{$v = 0.125 \times$} &  & \multicolumn{3}{c}{$v = 1 \times$} &  & \multicolumn{3}{c}{$v = 4 \times$} \\ \cmidrule(lr){2-4} \cmidrule(lr){6-8} \cmidrule(l){10-12} 
\multicolumn{1}{c}{\multirow{-2}{*}{Method}} & PSNR $\uparrow$ & SSIM $\uparrow$ & LPIPS $\downarrow$ &  & PSNR $\uparrow$ & SSIM $\uparrow$ & LPIPS $\downarrow$ &  & PSNR $\uparrow$ & SSIM $\uparrow$ & LPIPS $\downarrow$ \\ \midrule
E2VID $+$ NeRF & 18.58 & 0.849 & 0.259 &  & 18.85 & 0.839 & 0.278 &  & 17.82 & 0.804 & 0.328 \\
Robust \textit{e}-NeRF & 28.31 & 0.943 & 0.050 &  & 26.11 & 0.924 & 0.074 &  & 22.18 & 0.861 & 0.122 \\
\rowcolor[HTML]{F3F3F3} 
Deblur \textit{e}-NeRF$^\dag$ & \textbf{28.71} & \textbf{0.948} & \textbf{0.048} &  & \textbf{28.41} & \textbf{0.947} & \textbf{0.049} &  & \textbf{27.48} & \textbf{0.939} & \textbf{0.061} \\ \bottomrule
\end{tabular}

\end{table*}

\begin{table*}[t!]
\fontsize{7.5}{9}
\selectfont

\centering
\caption{Effect of scene illuminance. $^\dag$Trained with $\nicefrac{1}{8} \times$ the batch size of baselines.}
\label{tab:illuminance}

\begin{tabular}{@{}lccclccclccc@{}}
\toprule
\multicolumn{1}{c}{} & \multicolumn{3}{c}{$E_\mathit{sc} = 100 \ 000 \mathit{lux}$} &  & \multicolumn{3}{c}{$E_\mathit{sc} = 1 \ 000 \mathit{lux}$} &  & \multicolumn{3}{c}{$E_\mathit{sc} = 10 \mathit{lux}$} \\ \cmidrule(lr){2-4} \cmidrule(lr){6-8} \cmidrule(l){10-12} 
\multicolumn{1}{c}{\multirow{-2}{*}{Method}} & PSNR $\uparrow$ & SSIM $\uparrow$ & LPIPS $\downarrow$ &  & PSNR $\uparrow$ & SSIM $\uparrow$ & LPIPS $\downarrow$ &  & PSNR $\uparrow$ & SSIM $\uparrow$ & LPIPS $\downarrow$ \\ \midrule
E2VID $+$ NeRF & 19.27 & 0.846 & 0.268 &  & 18.85 & 0.839 & 0.278 &  & 17.24 & 0.804 & 0.354 \\
Robust \textit{e}-NeRF & 27.62 & 0.942 & 0.055 &  & 26.11 & 0.924 & 0.074 &  & 22.72 & 0.870 & 0.129 \\
\rowcolor[HTML]{F3F3F3} 
Deblur \textit{e}-NeRF$^\dag$ & \textbf{28.73} & \textbf{0.948} & \textbf{0.047} &  & \textbf{28.41} & \textbf{0.947} & \textbf{0.049} &  & \textbf{28.62} & \textbf{0.935} & \textbf{0.059} \\ \bottomrule
\end{tabular}

\end{table*}

\begin{table*}[t!]

\fontsize{5.5}{6.36}
\selectfont

\centering
\caption{Collective effect of camera speed and scene illuminance. $^\dag$Trained with $\nicefrac{1}{8} \times$ the batch size of baselines.}
\label{tab:collective}

\begin{tabular}{@{}lccccclccclccc@{}}
\toprule
\multicolumn{1}{c}{} &  &  & \multicolumn{3}{c}{$v= 0.125 \times, E_\mathit{sc} = 100 \ 000 \mathit{lux}$} &  & \multicolumn{3}{c}{$v= 1 \times, E_\mathit{sc} = 1 \ 000 \mathit{lux}$} &  & \multicolumn{3}{c}{$v= 4 \times, E_\mathit{sc} = 10 \mathit{lux}$} \\ \cmidrule(lr){4-6} \cmidrule(lr){8-10} \cmidrule(l){12-14} 
\multicolumn{1}{c}{\multirow{-2}{*}{Method}} & \multirow{-2}{*}{\begin{tabular}[c]{@{}c@{}}Opt.\\ $C_p$ \& $\tau$\end{tabular}} & \multirow{-2}{*}{\begin{tabular}[c]{@{}c@{}}Opt.\\ $\Omega$\end{tabular}} & PSNR $\uparrow$ & SSIM $\uparrow$ & LPIPS $\downarrow$ &  & PSNR $\uparrow$ & SSIM $\uparrow$ & LPIPS $\downarrow$ &  & PSNR $\uparrow$ & SSIM $\uparrow$ & LPIPS $\downarrow$ \\ \midrule
E2VID $+$ NeRF & $-$ & $-$ & 19.19 & 0.844 & 0.281 &  & 18.85 & 0.839 & 0.278 &  & 15.37 & 0.799 & 0.436 \\
 & $\times$ & $-$ & 28.27 & 0.944 & 0.057 &  & 26.11 & 0.924 & 0.074 &  & 18.42 & 0.814 & 0.255 \\
\multirow{-2}{*}{Robust \textit{e}-NeRF} & $\checkmark$ & $-$ & 28.28 & 0.944 & 0.051 &  & 26.31 & 0.923 & 0.075 &  & 18.51 & 0.812 & 0.254 \\
\rowcolor[HTML]{F3F3F3} 
\cellcolor[HTML]{F3F3F3} & $\times$ & $\times$ & \textbf{29.00} & \textbf{0.950} & \textbf{0.043} &  & \textbf{28.41} & \textbf{0.947} & \textbf{0.049} &  & \textbf{26.15} & \textbf{0.904} & \textbf{0.134} \\
\rowcolor[HTML]{F3F3F3} 
\multirow{-2}{*}{\cellcolor[HTML]{F3F3F3}Deblur \textit{e}-NeRF$^\dag$} & $\times$ & $\checkmark$ & 28.19 & 0.943 & 0.046 &  & 26.07 & 0.930 & 0.067 &  & 25.59 & 0.896 & 0.156 \\ \bottomrule
\end{tabular}

\end{table*}

\bfPar{Effect of Camera Speed.}
To investigate the effect of event motion blur due to high-speed camera motion, we evaluate all methods on 3 sets of sequences simulated with camera speeds $v$ that are $0.125 \times$, $1 \times$ \& $4 \times$ of the default setting, respectively. As event motion blur may lead to a significant ``loss or introduction of events'' (\cref{sec:intro}), we also quantify the average number of events relative to that of its corresponding blur-free sequence, across all sequences in the set. This translates to $93.36 \%$, $100.95 \%$ \& $95.21 \%$ for $v = 0.125 \times, 1 \times$ \& $4 \times$, respectively.

The quantitative results in \cref{tab:speed} clearly underscores the significance of incorporating a pixel bandwidth model, as our method significantly outperforms all baselines, especially under high-speed motion, despite being trained with $\nicefrac{1}{8} \times$ the batch size of baselines. The results also display our robustness to event motion blur, as our performance remains relatively unperturbed under varying camera speeds, \& remains close to our upper bound performance. Qualitative results in \cref{fig:qualitative} further validate our effectiveness in reconstructing a blur-minimal NeRF, as our method is free from artifacts such as floaters and double edges.

\bfPar{Effect of Scene Illuminance.}
To assess the effect of event motion blur due to poor scene illumination, we benchmark all methods on 3 sets of sequences simulated with scene illuminance $E_\mathit{sc}$ of $100 \ 000$, $1 \ 000$ \& $10 \mathit{lux}$, which correspond to sunlight, office light \& street light, respectively \cite{delbruck1995_phototransduction}. Furthermore, the pixel bandwidth spans around $1200$--$4000 \mathit{Hz}$, $50$--$2500 \mathit{Hz}$ \& $20$--$170 \mathit{Hz}$, respectively, depending on the incident radiance. They also have $82.49 \%$, $100.95 \%$ \& $17.22 \%$ the number of events of the blur-free set, respectively. The quantitative results given in \cref{tab:illuminance} undoubtedly verifies the importance of modeling event motion blur, as our method once again outperform other works, while remaining close to the upper bound performance. It also reveals our astonishing robustness to blur-induced ``loss of events'', particularly under $E_\mathit{sc} = 10 \mathit{lux}$.

\bfPar{Collective Effect.}
To assess the effect of event motion blur due to both high speed and low light, which resemble challenging real-world conditions, we benchmark all methods on 3 sets of sequences with different difficulty levels: \textit{easy} ($v= 0.125 \times, E_\mathit{sc} = 100 \ 000 \mathit{lux}$), \textit{medium} ($v= 1 \times, E_\mathit{sc} = 1 \ 000 \mathit{lux}$) and \textit{hard} ($v= 4 \times, E_\mathit{sc} = 10 \mathit{lux}$), which have $93.36 \%$, $100.95 \%$ and $7.14 \%$ the number of events of the blur-free set, respectively. We also evaluate Robust \textit{e}-NeRF with jointly optimized contrast thresholds $C_p$ \& refractory period $\tau$ to validate the importance of modeling event motion blur. We benchmark our method with jointly optimized pixel bandwidth model parameters $\Omega$, poorly initialized to $4\times$ their true value, to assess the robustness of joint optimization, thus auto-calibration.

The quantitative and qualitative results given in \cref{tab:collective} and  \cref{fig:qualitative}, respectively, generally reflect the results of the 2 previous experiments, thus a similar conclusion can be drawn. In addition, $C_p$ \& $\tau$ clearly cannot compensate for the lack of a pixel bandwidth model, as their joint optimization has virtually no effect on Robust \textit{e}-NeRF. The quantitative results also reveal the feasibility of jointly optimizing $\Omega$, especially under severe motion blur. While our performance deteriorates as the difficulty increases, which suggests a limit to our robustness to event motion blur, we show in the supplement that increasing our batch size to that of the baselines significantly improves our performance, hence robustness.

\subsection{Real Experiments}
\label{sec:exp:real}

To account for unknown event camera intrinsic parameters, we train Robust \textit{e}-NeRF and our method with jointly optimized contrast thresholds and refractory period. We also train our method with jointly optimized pixel bandwidth model parameters, initialized from DVS128 \cite{lichtsteiner2008_128} fast biases (provided in jAER \cite{delbruck2008_jaer}), using the same batch size as the baselines. While the 2 sequences only occasionally involve fast camera motion under moderate office lighting, the quantitative results reported in \cref{tab:real} demonstrate our superior performance. This is also supported qualitatively in \cref{fig:qualitative}, where the table \& walls are visibly more uniform, and less blooming artifacts are observed around objects, compared to Robust e-NeRF.

\section{Conclusion}
\label{sec:conclusion}

In this paper, we introduce Deblur \textit{e}-NeRF, a novel method to directly and effectively reconstruct blur-minimal NeRFs from motion-blurred events, generated under high-speed or low-light conditions. The core component of this work is a physically-accurate pixel bandwidth model that accounts for event motion blur. We also propose a threshold-normalized total variation loss to better regularize large textureless patches. Despite its accomplishments, Deblur \textit{e}-NeRF still inherits the limitations of Robust \textit{e}-NeRF and other works, \eg assumption of known camera trajectory. Moreover, since the synthesis of motion-blurred effective log-radiance at a given timestamp requires multiple past and present samples of effective log-radiance, given by NeRF renders, our reconstruction also incurs a higher computational and memory cost. Joint optimization of pixel bandwidth model parameters is also sometimes unstable. We leave these as future work.

\bfPar{Acknowledgements.}
This research / project is supported by the National Research Foundation,
Singapore, under its NRF-Investigatorship Programme (Award ID. NRF-NRFI09-0008), and the Tier 1
grant T1-251RES2305 from the Singapore Ministry of Education.

\bibliographystyle{splncs04}
\bibliography{references}

\begin{thebibliography}{10}
\providecommand{\url}[1]{\texttt{#1}}
\providecommand{\urlprefix}{URL }
\providecommand{\doi}[1]{https://doi.org/#1}

\bibitem{cannici2024_ev_deblurnerf}
Cannici, M., Scaramuzza, D.: Mitigating motion blur in neural radiance fields with events and frames. In: Proceedings of the IEEE/CVF Conference on Computer Vision and Pattern Recognition (CVPR) (2024)

\bibitem{cho2009_fast_motion_deblurring}
Cho, S., Lee, S.: Fast motion deblurring. ACM Trans. Graph.  (2009)

\bibitem{delbruck1995_phototransduction}
Delbruck, T., Mead, C.A.: Analog {VLSI} {Phototransduction} by continuous-time, adaptive, logarithmic photoreceptor circuits. Vision Chips: Implementing vision algorithms with analog VLSI circuits  (1995)

\bibitem{delbruck2008_jaer}
Delbruck, T.: Frame-free dynamic digital vision. In: Proceedings of Intl. Symp. on Secure-Life Electronics, Advanced Electronics for Quality Life and Society (2008)

\bibitem{delbruck1993_thesis}
Delbruck, T.: Investigations of analog {VLSI} visual transduction and motion processing. Doctoral {Thesis}, California Institute of Technology (1993)

\bibitem{franklin1998_digital_control}
Franklin, G., Powell, J., Workman, M.: Digital Control of Dynamic Systems. Addison-Wesley series in electrical and computer engineering: Control engineering, Addison-Wesley (1998)

\bibitem{gallego2020_event_survey}
Gallego, G., Delbr{\"u}ck, T., Orchard, G., Bartolozzi, C., Taba, B., Censi, A., Leutenegger, S., Davison, A.J., Conradt, J., Daniilidis, K., Scaramuzza, D.: Event-based vision: A survey. IEEE Transactions on Pattern Analysis and Machine Intelligence  (2020)

\bibitem{graca2021_unraveling}
Graca, R., Delbruck, T.: Unraveling the paradox of intensity-dependent dvs pixel noise. In: {International} {Image} {Sensor} {Workshop} ({IISW}) (2021)

\bibitem{graca2023_optimal_biasing}
Graca, R., McReynolds, B., Delbruck, T.: Optimal biasing and physical limits of dvs event noise. In: {International} {Image} {Sensor} {Workshop} ({IISW}) (2023)

\bibitem{graca2023_shining}
Graça, R., McReynolds, B., Delbruck, T.: Shining light on the dvs pixel: A tutorial and discussion about biasing and optimization. In: IEEE/CVF Conference on Computer Vision and Pattern Recognition Workshops (CVPRW) (2023)

\bibitem{hidalgo2022_eds}
Hidalgo-Carrió, J., Gallego, G., Scaramuzza, D.: Event-{Aided} {Direct} {Sparse} {Odometry}. In: Proceedings of the {IEEE}/{CVF} {Conference} on {Computer} {Vision} and {Pattern} {Recognition} ({CVPR}) (2022)

\bibitem{hu2021_v2e}
Hu, Y., Liu, S.C., Delbruck, T.: v2e: {From} {Video} {Frames} to {Realistic} {DVS} {Events}. In: Proceedings of the {IEEE}/{CVF} {Conference} on {Computer} {Vision} and {Pattern} {Recognition} ({CVPR}) {Workshops} (2021)

\bibitem{hwang2022_evnerf}
Hwang, I., Kim, J., Kim, Y.M.: Ev-nerf: Event based neural radiance field. In: IEEE/CVF Winter Conference on Applications of Computer Vision (WACV) (2023)

\bibitem{jiang2020_learning_event}
Jiang, Z., Zhang, Y., Zou, D., Ren, J., Lv, J., Liu, Y.: Learning event-based motion deblurring. In: Proceedings of the IEEE/CVF Conference on Computer Vision and Pattern Recognition (CVPR) (2020)

\bibitem{joubert2021_icns}
Joubert, D., Marcireau, A., Ralph, N., Jolley, A., van Schaik, A., Cohen, G.: Event {Camera} {Simulator} {Improvements} via {Characterized} {Parameters}. Frontiers in Neuroscience  (2021)

\bibitem{kato2020_dr_survey}
Kato, H., Beker, D., Morariu, M., Ando, T., Matsuoka, T., Kehl, W., Gaidon, A.: {Differentiable Rendering: A Survey} (2020), arXiv:2006.12057 [cs]

\bibitem{kingma2015_adam}
Kingma, D.P., Ba, J.: Adam: {A} method for stochastic optimization. In: 3rd International Conference on Learning Representations, {ICLR} 2015, San Diego, CA, USA, May 7-9, 2015, Conference Track Proceedings (2015)

\bibitem{klenk2022_enerf}
Klenk, S., Koestler, L., Scaramuzza, D., Cremers, D.: E-nerf: Neural radiance fields from a moving event camera. IEEE Robotics and Automation Letters  (2023)

\bibitem{krishnan2009_fast_image_deconv}
Krishnan, D., Fergus, R.: Fast image deconvolution using hyper-laplacian priors. In: Proceedings of the 22nd International Conference on Neural Information Processing Systems (2009)

\bibitem{krishnan2011_blind_deconv}
Krishnan, D., Tay, T., Fergus, R.: Blind deconvolution using a normalized sparsity measure. In: CVPR 2011 (2011)

\bibitem{lee2023_dp-nerf}
Lee, D., Lee, M., Shin, C., Lee, S.: Dp-nerf: Deblurred neural radiance field with physical scene priors. In: 2023 IEEE/CVF Conference on Computer Vision and Pattern Recognition (CVPR) (2023)

\bibitem{lee2023_exblurf}
Lee, D., Oh, J., Rim, J., Cho, S., Lee, K.: Exblurf: Efficient radiance fields for extreme motion blurred images. In: 2023 IEEE/CVF International Conference on Computer Vision (ICCV) (2023)

\bibitem{li2023_nerfacc}
Li, R., Gao, H., Tancik, M., Kanazawa, A.: Nerfacc: Efficient sampling accelerates nerfs. In: Proceedings of the IEEE/CVF International Conference on Computer Vision (ICCV) (2023)

\bibitem{lichtsteiner2006_aer}
Lichtsteiner, P.: An {AER} temporal contrast vision sensor. Doctoral {Thesis}, ETH Zurich (2006)

\bibitem{lichtsteiner2008_128}
Lichtsteiner, P., Posch, C., Delbruck, T.: A 128 $\times$ 128 120 {dB} 15 $\mu$s latency asynchronous temporal contrast vision sensor. IEEE Journal of Solid-State Circuits  (2008)

\bibitem{lin2020_learning_event_driven}
Lin, S., Zhang, J., Pan, J., Jiang, Z., Zou, D., Wang, Y., Chen, J., Ren, J.: Learning event-driven video deblurring and interpolation. In: Computer Vision -- ECCV 2020 (2020)

\bibitem{liu2024_nernet}
Liu, H., Peng, S., Zhu, L., Chang, Y., Zhou, H., Yan, L.: Seeing motion at nighttime with an event camera. In: Proceedings of the IEEE/CVF Conference on Computer Vision and Pattern Recognition (CVPR) (2024)

\bibitem{low2023_robust-e-nerf}
Low, W.F., Lee, G.H.: Robust e-nerf: Nerf from sparse \& noisy events under non-uniform motion. In: Proceedings of the IEEE/CVF International Conference on Computer Vision (ICCV) (2023)

\bibitem{ma2022_deblur-nerf}
Ma, L., Li, X., Liao, J., Zhang, Q., Wang, X., Wang, J., Sander, P.V.: Deblur-{NeRF}: {Neural} {Radiance} {Fields} {From} {Blurry} {Images}. In: Proceedings of the {IEEE}/{CVF} {Conference} on {Computer} {Vision} and {Pattern} {Recognition} ({CVPR}) (2022)

\bibitem{ma2023_de-nerf}
Ma, Q., Paudel, D.P., Chhatkuli, A., Van~Gool, L.: Deformable neural radiance fields using rgb and event cameras. In: Proceedings of the IEEE/CVF International Conference on Computer Vision (ICCV) (2023)

\bibitem{mcreynolds2022_experimental}
McReynolds, B.J., Graca, R.P., Delbruck, T.: Experimental methods to predict dynamic vision sensor event camera performance. Optical Engineering  (2022)

\bibitem{mildenhall2020_nerf}
Mildenhall, B., Srinivasan, P.P., Tancik, M., Barron, J.T., Ramamoorthi, R., Ng, R.: {NeRF: Representing Scenes as Neural Radiance Fields for View Synthesis}. In: European Conference on Computer Vision (ECCV) 2020 (2020)

\bibitem{muller2022_instant_ngp}
Müller, T., Evans, A., Schied, C., Keller, A.: Instant {Neural} {Graphics} {Primitives} with a {Multiresolution} {Hash} {Encoding}. ACM Transactions on Graphics  (2022)

\bibitem{nozaki2017_temperature}
Nozaki, Y., Delbruck, T.: Temperature and {Parasitic} {Photocurrent} {Effects} in {Dynamic} {Vision} {Sensors}. IEEE Transactions on Electron Devices  (2017)

\bibitem{park2017_joint_est}
Park, H., Mu~Lee, K.: Joint estimation of camera pose, depth, deblurring, and super-resolution from a blurred image sequence. In: Proceedings of the IEEE International Conference on Computer Vision (ICCV) (2017)

\bibitem{perrone2014_total_variation}
Perrone, D., Favaro, P.: Total variation blind deconvolution: The devil is in the details. In: Proceedings of the IEEE Conference on Computer Vision and Pattern Recognition (CVPR) (2014)

\bibitem{potmesil1983_motion_blur}
Potmesil, M., Chakravarty, I.: Modeling motion blur in computer-generated images. SIGGRAPH Comput. Graph.  (1983)

\bibitem{qi2023_e2nerf}
Qi, Y., Zhu, L., Zhang, Y., Li, J.: E2nerf: Event enhanced neural radiance fields from blurry images. In: Proceedings of the IEEE/CVF International Conference on Computer Vision (ICCV) (2023)

\bibitem{rebecq2018_esim}
Rebecq, H., Gehrig, D., Scaramuzza, D.: {ESIM}: an {Open} {Event} {Camera} {Simulator}. In: Proceedings of {The} 2nd {Conference} on {Robot} {Learning}, {PMLR} (2018)

\bibitem{rebecq2021_e2vid}
Rebecq, H., Ranftl, R., Koltun, V., Scaramuzza, D.: High speed and high dynamic range video with an event camera. IEEE Transactions on Pattern Analysis and Machine Intelligence  (2021)

\bibitem{rudnev2022_eventnerf}
Rudnev, V., Elgharib, M., Theobalt, C., Golyanik, V.: Eventnerf: Neural radiance fields from a single colour event camera (2022), arXiv:2206.11896 [cs]

\bibitem{shan2008_high_quality}
Shan, Q., Jia, J., Agarwala, A.: High-quality motion deblurring from a single image. ACM Trans. Graph.  (2008)

\bibitem{son2021_pvdnet}
Son, H., Lee, J., Lee, J., Cho, S., Lee, S.: Recurrent video deblurring with blur-invariant motion estimation and pixel volumes. ACM Trans. Graph.  (2021)

\bibitem{tao2018_srn-deblurnet}
Tao, X., Gao, H., Shen, X., Wang, J., Jia, J.: Scale-recurrent network for deep image deblurring. In: 2018 IEEE/CVF Conference on Computer Vision and Pattern Recognition (2018)

\bibitem{wang2023_pypose}
Wang, C., Gao, D., Xu, K., Geng, J., Hu, Y., Qiu, Y., Li, B., Yang, F., Moon, B., Pandey, A., Aryan, Xu, J., Wu, T., He, H., Huang, D., Ren, Z., Zhao, S., Fu, T., Reddy, P., Lin, X., Wang, W., Shi, J., Talak, R., Cao, K., Du, Y., Wang, H., Yu, H., Wang, S., Chen, S., Kashyap, A., Bandaru, R., Dantu, K., Wu, J., Xie, L., Carlone, L., Hutter, M., Scherer, S.: {PyPose}: A library for robot learning with physics-based optimization. In: IEEE/CVF Conference on Computer Vision and Pattern Recognition (CVPR) (2023)

\bibitem{wang2023_bad-nerf}
Wang, P., Zhao, L., Ma, R., Liu, P.: Bad-nerf: Bundle adjusted deblur neural radiance fields. In: 2023 IEEE/CVF Conference on Computer Vision and Pattern Recognition (CVPR) (2023)

\bibitem{wang2022_uformer}
Wang, Z., Cun, X., Bao, J., Zhou, W., Liu, J., Li, H.: Uformer: A general u-shaped transformer for image restoration. In: Proceedings of the IEEE/CVF Conference on Computer Vision and Pattern Recognition (CVPR) (2022)

\bibitem{zhou2004_ssim}
Wang, Z., Bovik, A., Sheikh, H., Simoncelli, E.: Image quality assessment: from error visibility to structural similarity. IEEE Transactions on Image Processing  (2004)

\bibitem{xu2021_motion_deblurring}
Xu, F., Yu, L., Wang, B., Yang, W., Xia, G.S., Jia, X., Qiao, Z., Liu, J.: Motion deblurring with real events. In: Proceedings of the IEEE/CVF International Conference on Computer Vision (ICCV). pp. 2583--2592 (October 2021)

\bibitem{xu2010_two_phase}
Xu, L., Jia, J.: Two-phase kernel estimation for robust motion deblurring. In: Proceedings of the 11th European Conference on Computer Vision: Part I (2010)

\bibitem{xu2013_unnatural_l0}
Xu, L., Zheng, S., Jia, J.: Unnatural l0 sparse representation for natural image deblurring. In: Proceedings of the IEEE Conference on Computer Vision and Pattern Recognition (CVPR) (2013)

\bibitem{yang2024_latency}
Yang, Y., Liang, J., Yu, B., Chen, Y., Ren, J.S., Shi, B.: Latency correction for event-guided deblurring and frame interpolation. In: Proceedings of the IEEE/CVF Conference on Computer Vision and Pattern Recognition (CVPR) (2024)

\bibitem{zamir2022_restomer}
Zamir, S.W., Arora, A., Khan, S., Hayat, M., Khan, F.S., Yang, M.H.: Restormer: Efficient transformer for high-resolution image restoration. In: Proceedings of the IEEE/CVF Conference on Computer Vision and Pattern Recognition (CVPR) (2022)

\bibitem{zamir2021_mprnet}
Zamir, S.W., Arora, A., Khan, S., Hayat, M., Khan, F.S., Yang, M.H., Shao, L.: Multi-stage progressive image restoration. In: Proceedings of the IEEE/CVF Conference on Computer Vision and Pattern Recognition (CVPR) (2021)

\bibitem{zhang2018_lpips}
Zhang, R., Isola, P., Efros, A.A., Shechtman, E., Wang, O.: The unreasonable effectiveness of deep features as a perceptual metric. In: 2018 IEEE/CVF Conference on Computer Vision and Pattern Recognition (CVPR) (2018)

\bibitem{zhang2022_unifying_motion}
Zhang, X., Yu, L.: Unifying motion deblurring and frame interpolation with events. In: Proceedings of the IEEE/CVF Conference on Computer Vision and Pattern Recognition (CVPR) (2022)

\end{thebibliography}

% ---- Supplementary Materials ----
\clearpage

\title{Supplementary Material for\\Deblur \textit{e}-NeRF: NeRF from Motion-Blurred Events under High-speed or Low-light Conditions}

\titlerunning{Deblur \textit{e}-NeRF}

\author{Weng Fei Low\orcidlink{0000-0001-7022-5713}\index{Low, Weng Fei} \and
Gim Hee Lee\orcidlink{0000-0002-1583-0475}\index{Lee, Gim Hee}}

\authorrunning{W. F. Low and G. H. Lee}

\institute{
The NUS Graduate School's Integrative Sciences and Engineering Programme (ISEP)\\
Institute of Data Science (IDS), National University of Singapore\\
Department of Computer Science, National University of Singapore\\
\email{\{wengfei.low, gimhee.lee\}@comp.nus.edu.sg}\\
\url{https://wengflow.github.io/deblur-e-nerf}
}

\maketitle

\appendix
\setcounter{equation}{13}
\setcounter{table}{5}
\setcounter{figure}{5}

\section{Logarithmic Photoreceptor}
\label{sec:photoreceptor}

\bfPar{Model.}
As mentioned in Sec.~3.1, we model the radiance-dependent band-limiting behavior of the logarithmic photoreceptor with the following unity-gain 2\textsuperscript{nd}-order Non-Linear Time-Invariant (NLTI) Low-Pass Filter (LPF) with input $u_p = \log L$, state
$\bm{x}_p = \begin{bsmallmatrix}
    \nicefrac{\partial \log L_\mathit{p}}{\partial t} &
    \log L_p
 \end{bsmallmatrix}^\top$ 
and output $y_p = \log L_p$:
\begin{equation}
    \begin{aligned}
        \dot{\bm{x}}_p ( t ) &= A_p \left( u_p(t) \right) \ \bm{x}_p ( t ) + B_p \left( u_p(t) \right) \ u_p(t) \\
        y_p (t) &= C_p \ \bm{x}_p ( t )
    \end{aligned} \ ,
    \label{eq:photoreceptor}
\end{equation}
\begin{flalign*}
    \text{where }
    A_p ( u ) =
    \begin{bmatrix}
        -2 \zeta ( u ) \omega_n ( u ) & -\omega_n^2 ( u ) \\
        1 & 0
    \end{bmatrix}
    , \ 
    B_p ( u ) =
    \begin{bmatrix}
        \omega_n^2 ( u ) \\
        0
    \end{bmatrix}
    , \ 
    C_p =
    \begin{bmatrix}
        0 & 1
    \end{bmatrix} . &
\end{flalign*}
The derivation of this model follows closely that of the small signal model for the original adaptive variant of the logarithmic photoreceptor circuit \cite{delbruck1993_thesis,delbruck1995_phototransduction}, but we account for the absence of an adaptive element in the circuit.

The radiance-dependent damping ratio $\zeta$ and natural angular frequency $\omega_n$ are, respectively, given by:
\begin{align}
    \zeta (u) &= \frac{ \tau_\mathit{out} + \tau_\mathit{in} (u) + \left( A_\mathit{amp} + 1 \right) \tau_\mathit{mil} (u) }{ 2 \sqrt{\tau_\mathit{out} \left( \tau_\mathit{in} (u) + \tau_\mathit{mil} (u) \right) \left( A_\mathit{loop} + 1 \right)} } \ , \label{eq:zeta}\\
    \omega_n (u) &= \sqrt{ \frac{A_\mathit{loop} + 1}{\tau_\mathit{out} \left( \tau_\mathit{in} (u) + \tau_\mathit{mil} (u) \right)} } \ , \label{eq:omega_n}
\end{align}
where $A_\mathit{amp}$ and $A_\mathit{loop}$ are the amplifier and total loop gains of the photoreceptor circuit, respectively, and $\tau_\mathit{out}$ is the time constant associated to the output node of the photoreceptor circuit and inversely proportional to the \textit{photoreceptor bias current} $I_\mathit{pr}$ (Fig.~3). Furthermore, $\tau_\mathit{in}$ and $\tau_\mathit{mil}$ are, respectively, the radiance-dependent time constants associated to the input node and \textit{Miller capacitance} of the photoreceptor circuit, given by:

\begin{align}
    \tau_\mathit{in} (u) &= \frac{C_\mathit{in} V_T}{\kappa \exp{u}} = \frac{C_\mathit{in} V_T}{\kappa L} \ , \label{eq:tau_in}\\
    \tau_\mathit{mil} (u) &= \frac{C_\mathit{mil} V_T}{\kappa \exp{u}} = \frac{C_\mathit{mil} V_T}{\kappa L} \ , \label{eq:tau_mil}
\end{align}
where $C_\mathit{in}$ and $C_\mathit{mil}$ are the (lumped) parasitic capacitance on the photodiode and Miller capacitance in the photoreceptor circuit, respectively, $V_T$ is the thermal voltage, and $\kappa$ is the signal photocurrent $I_p$ to incident radiance signal $L_\mathit{sig}$ ratio governed by the photodiode.

\bfPar{Behavior under Extreme Low Light.}
As $\tau_\mathit{out} \ll \tau_\mathit{in} + \tau_\mathit{mil}$ under extreme low light, the model described above reduces to a unity-gain 1\textsuperscript{st}-order NLTI LPF with input $u_{\hat{p}} = \log L$, state $x_{\hat{p}} =$ output $y_{\hat{p}} = \log L_p$:
\begin{equation}
    \begin{aligned}
        \dot{x}_{\hat{p}} ( t ) &= A_{\hat{p}} \left( u_{\hat{p}}(t) \right) \ x_{\hat{p}} ( t ) + B_{\hat{p}} \left( u_{\hat{p}}(t) \right) \ u_{\hat{p}}(t) \\
        y_{\hat{p}} (t) &= C_{\hat{p}} \ x_{\hat{p}} ( t )
    \end{aligned} \ ,
    \label{eq:photoreceptor_ell}
\end{equation}
where $A_{\hat{p}} ( u ) = -\omega_\mathit{c, \hat{p}} (u), B_{\hat{p}} ( u ) = \omega_\mathit{c, \hat{p}} (u)$ and $C_{\hat{p}} = 1$.

The cutoff angular frequency of this non-linear filter:
\begin{equation}
    \omega_\mathit{c, \hat{p}} (u) = \frac{A_\mathit{loop} + 1}{\tau_\mathit{in} (u) + \left( A_\mathit{amp} + 1 \right) \tau_\mathit{mil} (u)}
    \label{eq:omega_c_p-hat}
\end{equation}
is directly proportional to the effective radiance $L = L_\mathit{sig} + L_\mathit{dark}$. Nonetheless, it remains very much smaller than the radiance-independent cutoff angular frequencies of the source follower buffer $\omega_\mathit{c, sf}$ and differencing amplifier $\omega_\mathit{c, diff}$. Therefore, this rather simple 1\textsuperscript{st}-order model forms the dominant pole approximation of the full 4\textsuperscript{th}-order pixel bandwidth model under extreme low-light, which is relatively accurate. Furthermore, when $L (t) \approx L_\mathit{dark}$, we can further approximate this model with its linearized variant, which has a constant cutoff angular frequency of $\omega_\mathit{c, \hat{p}} ( \log L_\mathit{dark}) = \omega_\mathit{c, dom, min}$ (\cf Eq.~(11)).

\bfPar{Fundamental Limitations.}
The logarithmic photoreceptor 2\textsuperscript{nd}-order NLTI LPF is characterized by $A_\mathit{amp}$, $A_\mathit{loop}$, $\tau_\mathit{out}$, $\nicefrac{C_\mathit{in} V_T}{\kappa} = \tau_\mathit{in} L$ and $\nicefrac{C_\mathit{mil} V_T}{\kappa} = \tau_\mathit{mil} L$. When the unknown logarithmic photoreceptor model parameters are jointly optimized, the predicted effective radiance $\hat{\bm{L}}$ from the reconstructed NeRF is only accurate up a scale, since $\tau_\mathit{in}$ and $\tau_\mathit{mil}$ are invariant to the common scale of $L$, $\nicefrac{C_\mathit{in} V_T}{\kappa}$ and $\nicefrac{C_\mathit{mil} V_T}{\kappa}$. This further necessitates a translated-gamma correction (Sec.~3.4) of $\hat{\bm{L}}$ post-reconstruction.

\section{Event Simulator}
\label{sec:simulator}

\begin{figure}[t]
    \centering
    \includegraphics[width=1\linewidth]{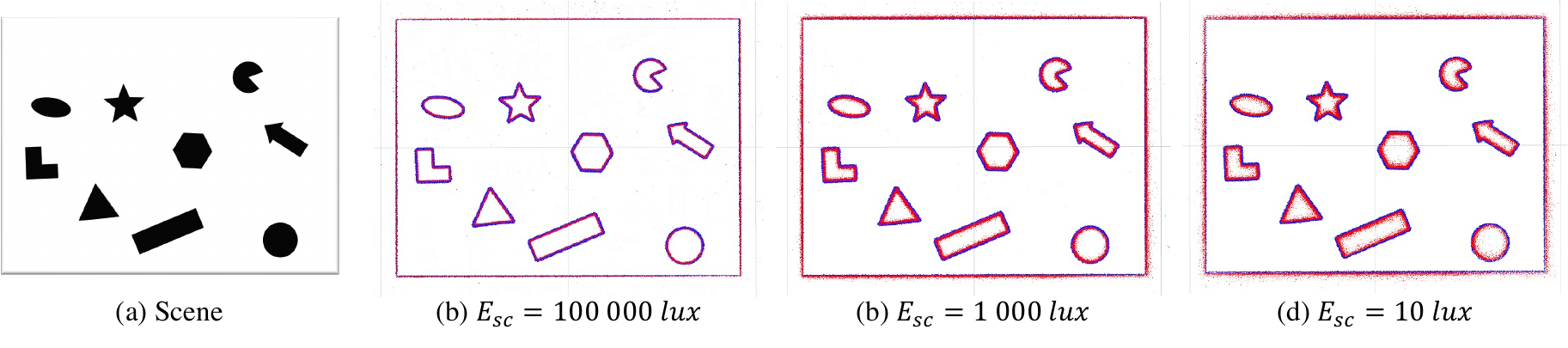}
    
    \caption{Our simulated events transformed to the scene plane.}
    \label{fig:simulated_events}
\end{figure}

As alluded in Secs.~3.2 and 4, our event simulator extends the improved ESIM \cite{rebecq2018_esim} event simulator introduced in Robust e-NeRF with the proposed pixel bandwidth model, particularly the discrete-time model given by Eq.~(8). We appropriately initialize the state of the 4\textsuperscript{th}-order NLTI LPF with the steady-state
$\bar{\bm{x}}[k_0] = 
\begin{bsmallmatrix}
    0 &
    u [k_0] &
    u [k_0] &
    u [k_0]
\end{bsmallmatrix}^\top$
on the initial input effective log-radiance $u[k_0] = \log L [k_0]$.

\cref{fig:simulated_events} depicts our simulated events on a simple planar scene of shapes under various scene illuminance $E_\mathit{sc}$. It can be observed that as the scene illumination improves, the events become more localized around the edges in the scene. Moreover, the spreading or blurring of negative events in red is more severe than that of positive events in blue. This happens because negative events involve a transition from a high to low effective log-radiance $\log L$, where the latter is associated to a low pixel bandwidth. All these observations validate the accuracy of our event simulator, as they conform to the expected behavior of an event sensor pixel.

\section{Implementation Details}
\label{sec:impl_details}

\bfPar{Deblur \textit{e}-NeRF.}
The implementation of our method is based on Robust \textit{e}-NeRF \cite{low2023_robust-e-nerf}. In particular, we adopt the same NeRF model architecture, parameterization of positive-to-negative contrast threshold ratio $\nicefrac{C_{+1}}{C_{-1}}$ and refractory period for joint optimization, NerfAcc \cite{li2023_nerfacc}/Instant-NGP \cite{muller2022_instant_ngp} parameters, training schedule, learning rates and constant-rate camera pose interpolation.

Nonetheless, since our method can theoretically reconstruct a NeRF with gamma-accurate predicted effective radiance $\hat{\bm{L}}$, particularly under unknown contrast thresholds, we also parameterize the mean contrast threshold $\bar{C} = \frac{1}{2} (C_{-1} + C_{+1})$, which defines the scale of the contrast thresholds, via \textit{SoftPlus} to ensure that it is always positive during its joint optimization. Such a parameterization of the contrast thresholds is optimal in the sense that the normalized predictions, which are $\nicefrac{\Delta \log \hat{L}_\mathit{blur}}{\bar{C}}$ for $\ell_{\mathit{diff}}$ and  $\nicefrac{\delta \log \hat{L}_\mathit{blur}}{\bar{C}}$ for $\ell_{\mathit{tv}}$, and normalized targets, which are $\nicefrac{p C_p}{\bar{C}}$ for $\ell_{\mathit{diff}}$ and  $0$ for $\ell_{\mathit{tv}}$, are invariant to $\nicefrac{C_{+1}}{C_{-1}}$ and $\bar{C}$, respectively.

Furthermore, we parameterize the pixel bandwidth model parameters as $A_\mathit{amp}^{-1}$, $A_\mathit{loop}^{-1}$, $\tau_\mathit{out}$, $\nicefrac{C_\mathit{mil} V_T}{\kappa} = \tau_\mathit{mil} L$, $\tau_\mathit{sf} = \omega_\mathit{c, sf}^{-1}$ and $\tau_\mathit{diff} = \omega_\mathit{c, diff}^{-1}$, which generally has values smaller than 1, via \textit{SoftPlus} as well for joint optimization. Note that we do not parameterize $\nicefrac{C_\mathit{in} V_T}{\kappa} = \tau_\mathit{in} L$, but keep it fixed at an arbitrary positive value, as the predicted effective radiance is only accurate up to a scale when pixel bandwidth model parameters are jointly optimized (\cref{sec:photoreceptor}). This helps to clamp down on this gauge freedom during joint optimization, and yields a minimal parameterization of $\tau_\mathit{in}$ and $\tau_\mathit{mil}$ up to an arbitrary scale. Care must be taken to ensure that the predefined $\nicefrac{C_\mathit{in} V_T}{\kappa}$ is larger than the minimum effective radiance $\epsilon = 0.001$ the NeRF model can output.

We adopt a sample size $k - k_0 + 1$ of $30$ for importance sampling of inputs $u = \log L$ in all experiments. Moreover, we sample the optimal input sample timestamps $T_i$ from the transformed exponential distribution given by Eq.~(11), but truncated in practice to a finite support of $\left( t_{k_0}, t_k \right]$ such that its cumulative probability is exactly $0.95$. The sampling is done using a variant of inverse transform sampling, where instead of uniformly sampling the interval $\left( 0, 1 \right]$ (and then applying the inverse cumulative distribution function), we directly take $k - k_0 + 1 = 30$ evenly-spaced samples in the same interval. This helps to prevent significant under/over-representation of inputs $u$ around certain time regions in the computation of the output $\bm{y} [k] $, due to randomness. Apart from that, since we assume $u$ is stationary prior the start of the event sequence, we assign input samples with timestamps prior the start to have the same value as the initial input.

Furthermore, we sample each subinterval $\left( t_\mathit{start}, t_\mathit{end} \right]$ between the interval $\left( t_\mathit{ref}, t_\mathit{curr} \right]$ for use in $\ell_{\mathit{tv}}$, by first sampling the length of the subinterval $t_\mathit{end} - t_\mathit{start}$ from a triangular distribution with a support of $\left[ 0, t_\mathit{curr} - t_\mathit{ref} \right)$ and a mode of $0$, then sampling $t_\mathit{start}$ from a uniform distribution with a support of $\left[ 0, \left( t_\mathit{curr} - t_\mathit{ref} \right) - \left( t_\mathit{end} - t_\mathit{start} \right) \right)$. Joint optimization of the pixel bandwidth model parameters is done with the same learning rate of 0.01 as the NeRF model parameters. Moreover, we train our method with loss weights of $\lambda_{\mathit{diff}} = 1$ and $\lambda_{\mathit{tv}} = 0.1$, as well as a batch size (defined relative to Robust \textit{e}-NeRF) of $2^{17} = 131 \ 072$, by default.

However, we observed that our loss values for the threshold-normalized difference loss $\ell_{\mathit{diff}}$, under the \textit{hard} setting ($v= 4 \times, E_\mathit{sc} = 10 \mathit{lux}$) in the collective effect synthetic experiment, is $\sim 100 \times$ smaller than that of other settings, but the loss values for threshold-normalized total variation loss $\ell_{\mathit{tv}}$ (Eq.~(12)) remains in the same order. This will cause the total training loss $\mathcal{L}$ (Eq.~(2), but with $\lambda_{\mathit{tv}} \ell_{\mathit{tv}} (\bm{e})$ instead of $\lambda_{\mathit{grad}} \ell_{\mathit{grad}} (\bm{e})$) to be inappropriately dominated by the regularization loss $\ell_{\mathit{tv}}$, instead of the primary reconstruction loss $\ell_{\mathit{diff}}$, if the default loss weights are used.

Thus, we adopt $\lambda_{\mathit{tv}} = 0.001$, which is $ 100 \times$ smaller than the default, under the \textit{hard} setting to rebalance both losses. Apart from that, we also adopt $\lambda_{\mathit{tv}} = 0.01$, which is $ 10 \times$ smaller than the default, under the $E_\mathit{sc} = 10 \mathit{lux}$ setting in the synthetic experiment studying the effect of scene illuminance, due to similar observations. As we employ the Adam \cite{kingma2015_adam} optimizer, which is invariant to diagonal rescaling of gradients hence loss, this is equivalent to a loss weight $\lambda_{\mathit{diff}}$ of $100 \times$ or $10 \times$ larger than that of the default, while maintaining $\lambda_{\mathit{tv}}$ at the default.

\bfPar{Baselines.}
We employ the official implementation of Robust \textit{e}-NeRF in our experiments. However, we adopt $\lambda_{\mathit{grad}} = 0.00001$ and $\lambda_{\mathit{grad}} = 0.0001$, which are $100 \times$ and $10 \times$ smaller than the default at $\lambda_{\mathit{grad}} = 0.001$, under the \textit{hard} ($v= 4 \times, E_\mathit{sc} = 10 \mathit{lux}$) and $E_\mathit{sc} = 10 \mathit{lux}$ settings in synthetic experiment, respectively, due to similar observations made in our method. Moreover, we implement E2VID $+$ NeRF according to how it is done for the experiments in Robust \textit{e}-NeRF.

\bfPar{Translated-Gamma Correction.}
To account for the time-varying sensor gain (\ie ISO) and exposure time of the captured reference images, particularly during evaluation, we additionally scale each correction with the known gain-exposure product of the corresponding reference image.

The optimal correction parameters $a, \bm{b}$ and $\bm{c}$ are optimized using the Levenberg-Marquardt algorithm with a \textit{Trust Region} strategy to determine the optimal damping factor at each iteration. We adopt the implementation provided by PyPose \cite{wang2023_pypose}, as well as its default hyperparameters. Furthermore, we appropriately initialize the optimization with $\bm{c} = \bm{0}$ and the solution of $a$ and $\bm{b}$ given by gamma correction (Eq.~(5)). The optimization is performed until the sum of squared correction errors has converged, up to a maximum of 20 iterations.

\section{Interpretation of Real Quantitative Results}
\label{sec:interpret}

Note that care must be taken when interpreting the quantitative results of the real experiments presented in Tab.~2 and \cref{tab:07_ziggy_and_fuzz_hdr}, since they are not truly indicative of the \textit{absolute} performance of all methods, but likely only indicative of their \textit{relative} performance. This is due to the fact that the target novel views, given by a separate standard camera, suffer from motion blur, rolling shutter artifacts, and saturation, as a result of a significantly smaller dynamic range compared to an event camera. Furthermore, the target novel views are not raw images that does not depend on the unknown \textit{Camera Response Function} (CRF), and are grayscale images converted from RGB images provided by the camera, which might not reflect the spectral sensitivity of the monochrome event camera.

\section{Additional Experiment Results}
\label{sec:add_results}

\begin{table}[t!]

\centering
\caption{Per-synthetic scene breakdown under the hard setting. $^\dag$Trained with $\nicefrac{1}{8} \times$ the batch size of baselines.}
\label{tab:breakdown}

\begin{tabular}{@{}lllccccccclc@{}}
\toprule
\multicolumn{1}{c}{} & \multicolumn{1}{c}{} &  & \multicolumn{7}{c}{Synthetic Scene} &  &  \\ \cmidrule(lr){4-10}
\multicolumn{1}{c}{\multirow{-2}{*}{Metric}} & \multicolumn{1}{c}{\multirow{-2}{*}{Method}} &  & Chair & Drums & Ficus & Hotdog & Lego & Materials & Mic &  & \multirow{-2}{*}{Mean} \\ \midrule
 & E2VID $+$ NeRF &  & 16.67 & 15.00 & 16.25 & 17.53 & 14.75 & 11.65 & 15.72 &  & 15.37 \\
 & Robust \textit{e}-NeRF &  & 21.64 & 17.41 & 21.80 & 15.05 & 18.28 & 15.68 & 19.11 &  & 18.42 \\
\multirow{-3}{*}{PSNR $\uparrow$} & \cellcolor[HTML]{F3F3F3}Deblur \textit{e}-NeRF$^\dag$ & \cellcolor[HTML]{F3F3F3} & \cellcolor[HTML]{F3F3F3}\textbf{27.39} & \cellcolor[HTML]{F3F3F3}\textbf{22.14} & \cellcolor[HTML]{F3F3F3}\textbf{29.10} & \cellcolor[HTML]{F3F3F3}\textbf{23.69} & \cellcolor[HTML]{F3F3F3}\textbf{27.69} & \cellcolor[HTML]{F3F3F3}\textbf{24.49} & \cellcolor[HTML]{F3F3F3}\textbf{28.53} & \cellcolor[HTML]{F3F3F3} & \cellcolor[HTML]{F3F3F3}\textbf{26.15} \\ \midrule
 & E2VID $+$ NeRF &  & 0.835 & 0.776 & 0.840 & 0.842 & 0.719 & 0.726 & 0.854 &  & 0.799 \\
 & Robust \textit{e}-NeRF &  & 0.836 & 0.758 & 0.864 & 0.849 & 0.754 & 0.772 & 0.862 &  & 0.814 \\
\multirow{-3}{*}{SSIM $\uparrow$} & \cellcolor[HTML]{F3F3F3}Deblur \textit{e}-NeRF$^\dag$ & \cellcolor[HTML]{F3F3F3} & \cellcolor[HTML]{F3F3F3}\textbf{0.902} & \cellcolor[HTML]{F3F3F3}\textbf{0.839} & \cellcolor[HTML]{F3F3F3}\textbf{0.944} & \cellcolor[HTML]{F3F3F3}\textbf{0.904} & \cellcolor[HTML]{F3F3F3}\textbf{0.896} & \cellcolor[HTML]{F3F3F3}\textbf{0.890} & \cellcolor[HTML]{F3F3F3}\textbf{0.951} & \cellcolor[HTML]{F3F3F3} & \cellcolor[HTML]{F3F3F3}\textbf{0.904} \\ \midrule
 & E2VID $+$ NeRF &  & 0.374 & 0.498 & 0.310 & 0.391 & 0.509 & 0.589 & 0.380 &  & 0.436 \\
 & Robust \textit{e}-NeRF &  & 0.216 & 0.336 & 0.146 & 0.287 & 0.279 & 0.295 & 0.228 &  & 0.255 \\
\multirow{-3}{*}{LPIPS $\downarrow$} & \cellcolor[HTML]{F3F3F3}Deblur \textit{e}-NeRF$^\dag$ & \cellcolor[HTML]{F3F3F3} & \cellcolor[HTML]{F3F3F3}\textbf{0.107} & \cellcolor[HTML]{F3F3F3}\textbf{0.231} & \cellcolor[HTML]{F3F3F3}\textbf{0.120} & \cellcolor[HTML]{F3F3F3}\textbf{0.168} & \cellcolor[HTML]{F3F3F3}\textbf{0.105} & \cellcolor[HTML]{F3F3F3}\textbf{0.116} & \cellcolor[HTML]{F3F3F3}\textbf{0.093} & \cellcolor[HTML]{F3F3F3} & \cellcolor[HTML]{F3F3F3}\textbf{0.134} \\ \bottomrule
\end{tabular}

\end{table}

\begin{figure}[t!]
    \centering
    \includegraphics[width=1\linewidth]{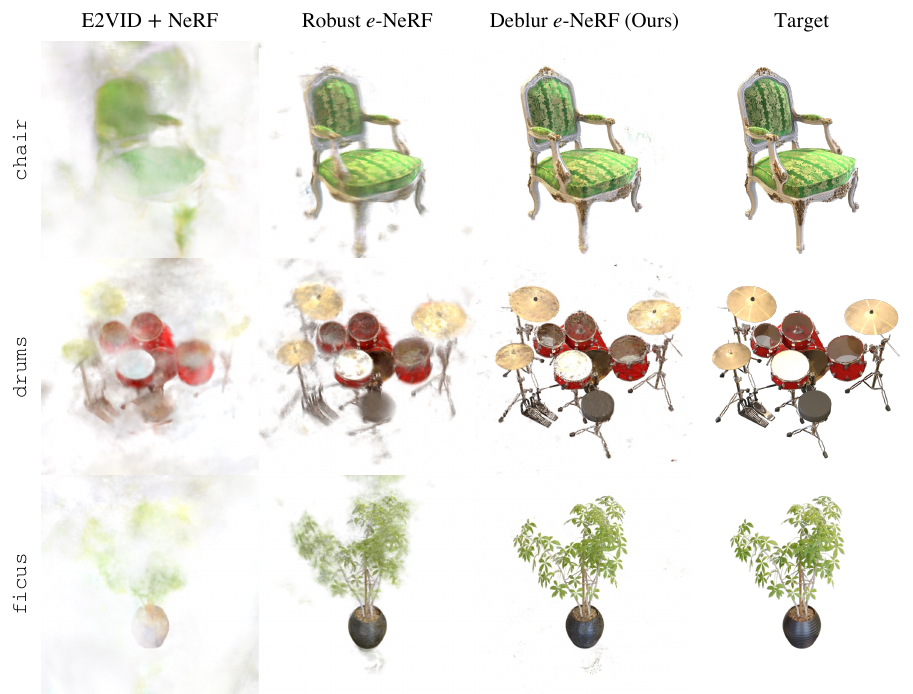}

    \caption{Synthesized novel views on \texttt{chair},  \texttt{drums} and \texttt{ficus} under the hard setting}
    \label{fig:breakdown1}
\end{figure}

\begin{figure}[t!]
    \centering
    \includegraphics[width=1\linewidth]{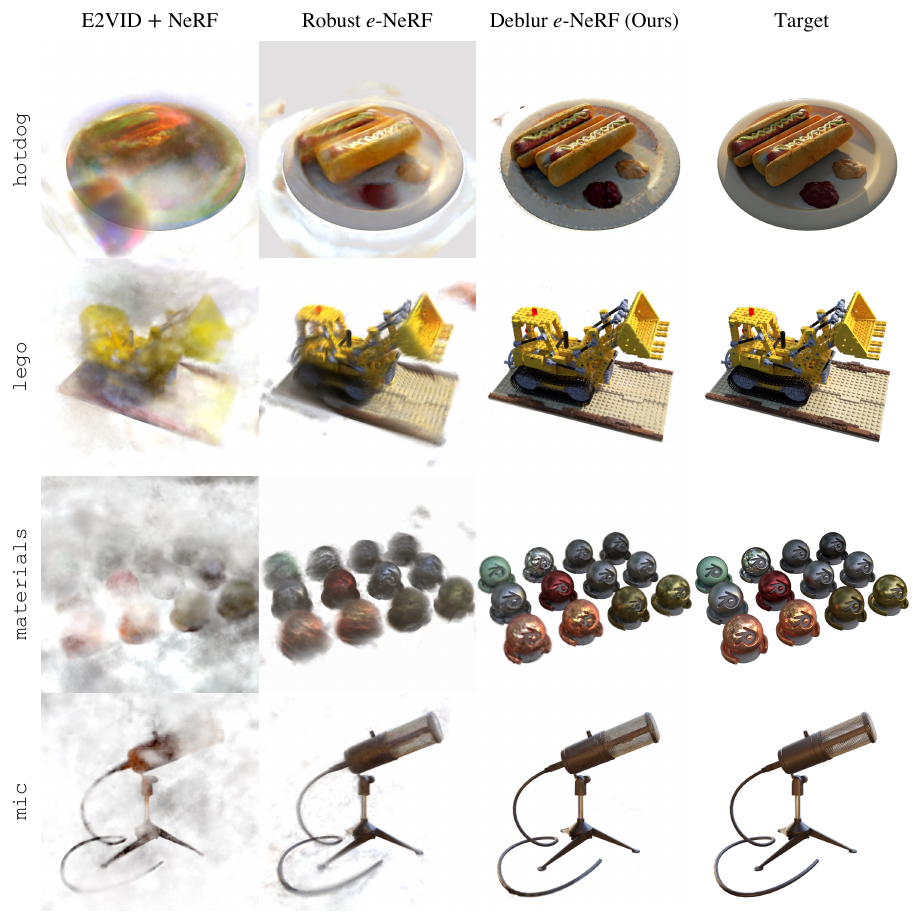}

    \caption{Synthesized novel views on \texttt{hotdog}, \texttt{lego},  \texttt{materials} and \texttt{mic} under the hard setting}
    \label{fig:breakdown2}
\end{figure}

\subsection{Per-Scene Breakdown}
\label{sec:add_results:per_scene}

\cref{tab:breakdown} and  \cref{fig:breakdown1,fig:breakdown2} show the quantitative and qualitative results of all methods, respectively, for all synthetic scene sequences simulated with the hard setting  ($v= 4 \times, E_\mathit{sc} = 10 \mathit{lux}$). The results clearly demonstrate our superior performance in reconstructing a blur-minimal NeRF from motion-blurred events.

\subsection{Ablation on Pixel Bandwidth Model}
\label{sec:add_results:pbm}

To further ascertain the role of the proposed pixel bandwidth model, we evaluate our method with and without the pixel bandwidth model incorporated, under the same settings as the synthetic experiment in studying the collective effect, without joint optimization of pixel bandwidth model parameters. The quantitative results given in \cref{tab:pbm} undoubtedly verifies the importance of the pixel bandwidth model in accounting for event motion blur.

\subsection{Effect of Reduced Batch Size}
\label{sec:add_results:batch_size}

To assess the true impact of training with a reduced batch size, we also benchmark our method with a batch size of $\nicefrac{1}{8} \times$ and $1 \times$ that of our baselines, under the same settings as the experiment in \cref{sec:add_results:pbm}, but only on the \texttt{lego} scene. The quantitative results reported in \cref{tab:batch_size} 
provide a glimpse into the true strength of our method, as significant improvements can be observed as the batch size increases to that of the baselines.

\subsection{Ablation on Input Sample Size}
\label{sec:add_results:sample_size}

We perform a cost-benefit analysis on the input sample size $k - k_0 + 1$ of our method on the \texttt{lego} scene under the hard setting ($v= 4 \times, E_\mathit{sc} = 10 \mathit{lux}$). The quantitative results presented in \cref{tab:sample_size} suggests that our default input sample size of $30$ strikes the best balance between cost and performance. Note that the computational and memory cost of our method is proportional to the input sample size, as alluded in Sec.~5, and an input sample size of $1$ is equivalent to having the pixel bandwidth model removed.

\begin{table}[t!]
\fontsize{7.5}{9}
\selectfont

\centering
\caption{Ablation on pixel bandwidth model}
\label{tab:pbm}

\begin{tabular}{@{}clccclccclccc@{}}
\toprule
 &  & \multicolumn{3}{c}{$v= 0.125 \times, E_\mathit{sc} = 100 \ 000 \mathit{lux}$} &  & \multicolumn{3}{c}{$v= 1 \times, E_\mathit{sc} = 1 \ 000 \mathit{lux}$} &  & \multicolumn{3}{c}{$v= 4 \times, E_\mathit{sc} = 10 \mathit{lux}$} \\ \cmidrule(lr){3-5} \cmidrule(lr){7-9} \cmidrule(l){11-13} 
\multirow{-2}{*}{\begin{tabular}[c]{@{}c@{}}Pixel Bandwidth\\ Model\end{tabular}} &  & PSNR $\uparrow$ & SSIM $\uparrow$ & LPIPS $\downarrow$ &  & PSNR $\uparrow$ & SSIM $\uparrow$ & LPIPS $\downarrow$ &  & PSNR $\uparrow$ & SSIM $\uparrow$ & LPIPS $\downarrow$ \\ \midrule
$\times$ &  & 28.75 & 0.948 & 0.048 &  & 26.98 & 0.934 & 0.061 &  & 18.31 & 0.822 & 0.245 \\
\rowcolor[HTML]{F3F3F3} 
$\checkmark$ &  & \textbf{29.00} & \textbf{0.950} & \textbf{0.043} &  & \textbf{28.41} & \textbf{0.947} & \textbf{0.049} &  & \textbf{26.15} & \textbf{0.904} & \cellcolor[HTML]{F3F3F3}\textbf{0.134} \\ \bottomrule
\end{tabular}

\end{table}

\begin{table}[t!]
\fontsize{7.5}{9}
\selectfont

\centering
\caption{Effect of reduced batch size on the \texttt{lego} scene}
\label{tab:batch_size}

\begin{tabular}{@{}clccclccclccc@{}}
\toprule
 &  & \multicolumn{3}{c}{$v= 0.125 \times, E_\mathit{sc} = 100 \ 000 \mathit{lux}$} &  & \multicolumn{3}{c}{$v= 1 \times, E_\mathit{sc} = 1 \ 000 \mathit{lux}$} &  & \multicolumn{3}{c}{$v= 4 \times, E_\mathit{sc} = 10 \mathit{lux}$} \\ \cmidrule(lr){3-5} \cmidrule(lr){7-9} \cmidrule(l){11-13} 
\multirow{-2}{*}{Batch Size, $\times$} &  & PSNR $\uparrow$ & SSIM $\uparrow$ & LPIPS $\downarrow$ &  & PSNR $\uparrow$ & SSIM $\uparrow$ & LPIPS $\downarrow$ &  & PSNR $\uparrow$ & SSIM $\uparrow$ & LPIPS $\downarrow$ \\ \midrule
\rowcolor[HTML]{F3F3F3} 
$\nicefrac{1}{8}$ &  & 29.44 & 0.940 & 0.045 &  & 28.42 & 0.938 & 0.048 &  & 27.69 & 0.896 & 0.105 \\
$1$ &  & \textbf{31.27} & \textbf{0.953} & \textbf{0.030} &  & \textbf{30.43} & \textbf{0.950} & \textbf{0.038} &  & \textbf{30.72} & \textbf{0.948} & \textbf{0.037} \\ \bottomrule
\end{tabular}

\end{table}

\begin{table}[t!]

\centering
\caption{Ablation of input sample size on \texttt{lego} under the hard setting}
\label{tab:sample_size}

\begin{tabular}{@{}cccc@{}}
\toprule
Input Sample Size & PSNR $\uparrow$ & SSIM $\uparrow$ & LPIPS $\downarrow$ \\ \midrule
1 & 18.46 & 0.765 & 0.273 \\
5 & 22.64 & 0.807 & 0.21 \\
15 & 26.41 & 0.875 & 0.125 \\
\rowcolor[HTML]{F3F3F3} 
30 & 27.69 & 0.896 & 0.105 \\
50 & 28.18 & 0.902 & 0.097 \\
75 & \textbf{28.21} & \textbf{0.903} & \textbf{0.096} \\ \bottomrule
\end{tabular}

\end{table}

\subsection{Results on \texttt{07\_ziggy\_and\_fuzz\_hdr}}
\label{sec:add_results:07_ziggy_and_fuzz_hdr}

\begin{figure}[t!]
    \centering
    \includegraphics[width=1\linewidth]{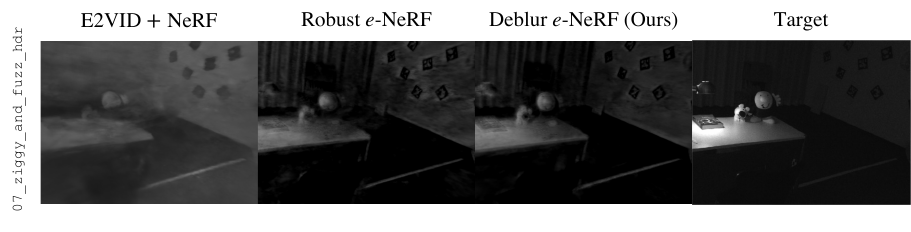}

    \caption{Synthesized novel views on the \texttt{07\_ziggy\_and\_fuzz\_hdr} scene}
    \label{fig:07_ziggy_and_fuzz_hdr}
\end{figure}

Apart from \texttt{08\_peanuts\_running} and \texttt{11\_all\_characters}, we also benchmark all methods on the \texttt{07\_ziggy\_and\_fuzz\_hdr} sequence from the EDS dataset, which involves a HDR scene with occasional high-speed camera motion. The quantitative and qualitative results given in \cref{tab:07_ziggy_and_fuzz_hdr} and \cref{fig:07_ziggy_and_fuzz_hdr} once again demonstrates our superior performance, as the objects on the table are clearly more well-defined and the table surface, wall and curtains are much smoother, while preserving details and color accuracy of the scene.

\subsection{Comparison with Image Blur-Aware Baselines}
\label{sec:add_results:img_blur_aware_baselines}

While motion blur in standard and event cameras are vastly different, and thus incomparable, we provide additional quantitative results of 2 other \textit{image} blur-aware baselines: E2VID $+$ MPRNet \cite{zamir2021_mprnet} (a seminal image deblurring method) $+$ NeRF and E2VID $+$ Deblur-NeRF \cite{ma2022_deblur-nerf} (a seminal NeRF with image blur model), for selected synthetic experiments (\ie upper bound performance and collective effect) and the real experiment, to make our experiments more complete. From the results reported in \cref{tab:img_blur_aware_baselines}, it is evident that the incorporation of an image blur or deblur model is unable to account for event motion blur, as the performance is virtually the same with or without it. This reinforces the importance for our physically-accurate pixel bandwidth model to account for event motion blur under arbitrary speed and lighting conditions.

\begin{table}[t!]

\centering
\caption{Quantitative results on the \texttt{07\_ziggy\_and\_fuzz\_hdr} scene}
\label{tab:07_ziggy_and_fuzz_hdr}

\begin{tabular}{@{}lccc@{}}
\toprule
\multicolumn{1}{c}{Method} & PSNR $\uparrow$ & SSIM $\uparrow$ & LPIPS $\downarrow$ \\ \midrule
E2VID $+$ NeRF & 14.96 & \textbf{0.691} & 0.556 \\
Robust \textit{e}-NeRF & 18.02 & 0.631 & 0.464 \\
\rowcolor[HTML]{F3F3F3} 
Deblur \textit{e}-NeRF & \textbf{18.47} & 0.648 & \textbf{0.440} \\ \bottomrule
\end{tabular}

\end{table}
\begin{table*}[t!]
\setlength{\tabcolsep}{1.9pt}
\fontsize{5.5}{6.6}
\selectfont

\centering
\caption{Comparison with image blur-aware baselines built upon E2VID.}
\label{tab:img_blur_aware_baselines}

\begin{tabular}{@{}lccclccclccc@{}}
\toprule
\multicolumn{1}{c}{\multirow{2}{*}{\begin{tabular}[c]{@{}c@{}}Simulation Settings\\ / Real Scene\end{tabular}}} & \multicolumn{3}{c}{E2VID $+$ NeRF} &  & \multicolumn{3}{c}{E2VID $+$ MPRNet $+$ NeRF} &  & \multicolumn{3}{c}{E2VID $+$ Deblur-NeRF} \\ \cmidrule(lr){2-4} \cmidrule(lr){6-8} \cmidrule(l){10-12} 
\multicolumn{1}{c}{} & PSNR $\uparrow$ & SSIM $\uparrow$ & LPIPS $\downarrow$ &  & PSNR $\uparrow$ & SSIM $\uparrow$ & LPIPS $\downarrow$ &  & PSNR $\uparrow$ & SSIM $\uparrow$ & LPIPS $\downarrow$ \\ \midrule
No event motion blur & 19.49 & 0.847 & 0.268 &  & 19.44 & \textbf{0.851} & \textbf{0.267} &  & \textbf{19.84} & 0.839 & 0.291 \\
$v= 0.125 \times, E_\mathit{sc} = 100 \ 000 \mathit{lux}$ & \textbf{19.19} & 0.844 & 0.281 &  & 19.18 & \textbf{0.849} & \textbf{0.260} &  & 19.15 & 0.841 & 0.288 \\
$v= 1 \times, E_\mathit{sc} = 1 \ 000 \mathit{lux}$ & 18.85 & 0.839 & 0.278 &  & \textbf{18.86} & \textbf{0.843} & \textbf{0.269} &  & 18.73 & 0.818 & 0.317 \\
$v= 4 \times, E_\mathit{sc} = 10 \mathit{lux}$ & 15.37 & \textbf{0.799} & \textbf{0.436} &  & \textbf{15.44} & 0.794 & 0.439 &  & 15.42 & 0.783 & 0.472 \\ \midrule
\texttt{07\_ziggy\_and\_fuzz\_hdr} & \textbf{14.96} & \textbf{0.691} & 0.556 &  & \textbf{14.96} & \textbf{0.691} & 0.552 &  & 14.85 & 0.680 & \textbf{0.504} \\
\texttt{08\_peanuts\_running} & 14.85 & \textbf{0.690} & 0.595 &  & 14.81 & \textbf{0.690} & 0.604 &  & \textbf{14.91} & 0.682 & \textbf{0.517} \\
\texttt{11\_all\_characters} & \textbf{13.12} & \textbf{0.695} & 0.627 &  & 13.10 & \textbf{0.695} & 0.624 &  & 12.95 & 0.689 & \textbf{0.576} \\ \bottomrule
\end{tabular}

\end{table*}

\end{document}